\documentclass[10pt,twocolumn,letterpaper]{article}

\usepackage[pagenumbers]{cvpr} 

\usepackage[dvipsnames]{xcolor}

\usepackage{algorithm}
\usepackage{algpseudocode}
\usepackage{multirow}
\usepackage{dsfont}
\usepackage{pifont}
\newcommand{\cmark}{\ding{51}}
\newcommand{\xmark}{\ding{55}}
\usepackage{arydshln}
\usepackage[subtle,mathdisplays=normal,wordspacing=normal]{savetrees}
\usepackage[accsupp]{axessibility} 
\usepackage{indentfirst}
\usepackage{mathtools}
\usepackage{xspace}

\newcommand{\mbd}{\mathbf{d}}

\newcommand{\mbm}{\mathbf{m}}

\newcommand{\mbr}{\mathbf{r}}

\newcommand{\mbt}{\mathbf{t}}

\newcommand{\mbx}{\mathbf{x}}

\newcommand{\mbz}{\mathbf{z}}

\newcommand{\mbD}{\mathbf{D}}

\newcommand{\mbF}{\mathbf{F}}

\newcommand{\mbH}{\mathbf{H}}

\newcommand{\mbJ}{\mathbf{J}}

\newcommand{\mbM}{\mathbf{M}}

\newcommand{\mbn}{\mathbf{n}}

\newcommand{\mbP}{\mathbf{P}}

\newcommand{\mbT}{\mathbf{T}}

\newcommand{\mbV}{\mathbf{V}}

\newcommand{\mbX}{\mathbf{X}}

\newcommand{\ignore}[1]{}

\makeatletter
\DeclareRobustCommand\onedot{\futurelet\@let@token\@onedot}
\def\@onedot{\ifx\@let@token.\else.\null\fi\xspace}
\def\eg{{e.g}\onedot} 
\def\ie{{i.e}\onedot} 
\makeatother

\newcommand\blfootnote[1]{%
  \begingroup
  \renewcommand\thefootnote{}\footnote{#1}%
  \addtocounter{footnote}{-1}%
  \endgroup
}
\definecolor{cvprblue}{rgb}{0.21,0.49,0.74}
\usepackage[pagebackref,breaklinks,colorlinks,citecolor=cvprblue]{hyperref}

\def\paperID{6073} 
\def\confName{CVPR}
\def\confYear{2024}

\title{Text2HOI: Text-guided 3D Motion Generation for Hand-Object Interaction}

\author{Junuk Cha\textsuperscript{1} \qquad Jihyeon Kim\textsuperscript{1,2\dag} \qquad Jae Shin Yoon\textsuperscript{3*} \qquad Seungryul Baek\textsuperscript{1*} \vspace{0.3em} \\
{\normalsize $^1$UNIST} \qquad
{\normalsize $^2$KETI} \qquad
{\normalsize $^3$Adobe Research}
}

\begin{document}

\twocolumn[{
\maketitle
\begin{center}
    \captionsetup{type=figure}
    \vspace{-6mm}
    \includegraphics[width=0.99\textwidth]{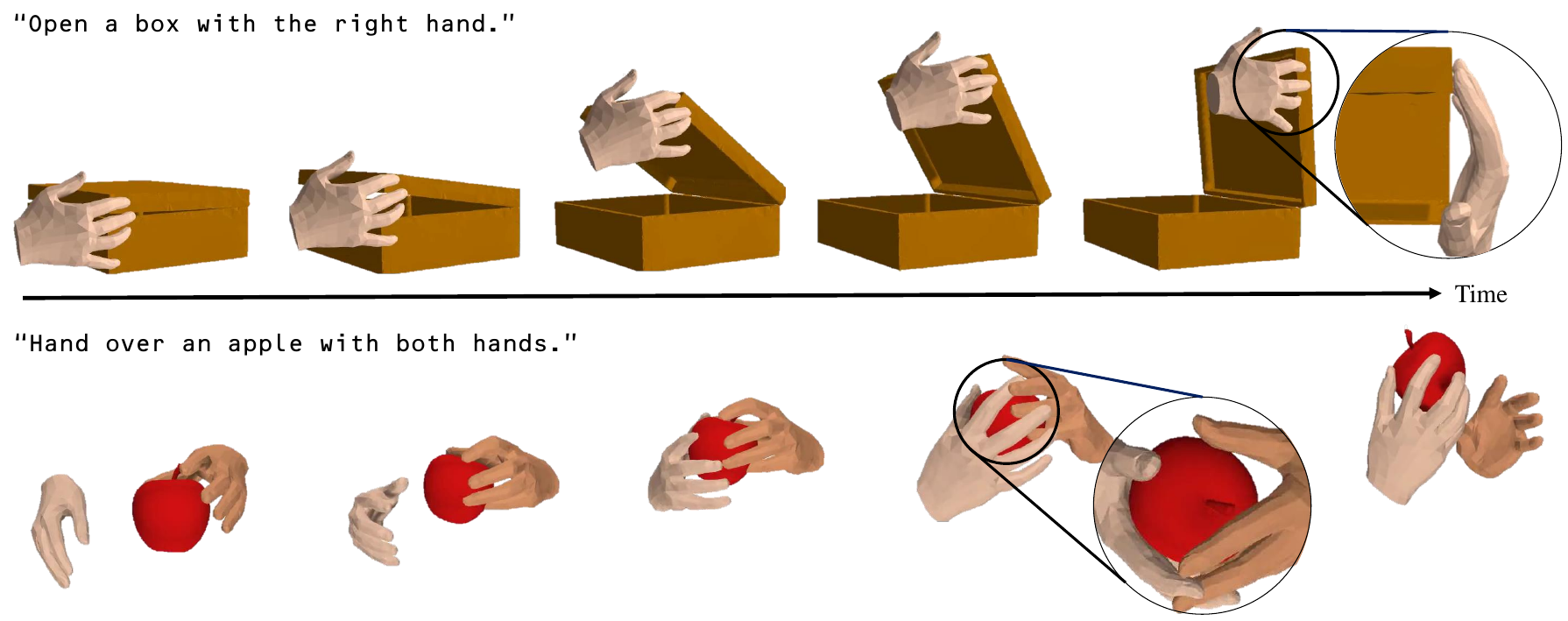}
    \vspace{-3mm}
    \captionof{figure}{Given a text and a canonical object mesh as prompts, we generate 3D motion for hand-object interaction without requiring object trajectory and initial hand pose. We represent the right hand with a light skin color and the left hand with a dark skin color. The articulation of a box in the first row is controlled by estimating an angle for the pre-defined axis of the box.}
    \label{fig:teaser}
\end{center}
}]
\blfootnote{This research was conducted when Jihyeon Kim was a graduate student (Master candidate) at UNIST$\dag$. Co-last authors$*$.}

\begin{abstract}
This paper introduces the first text-guided work for generating the sequence of hand-object interaction in 3D.
The main challenge arises from the lack of labeled data where existing ground-truth datasets are nowhere near generalizable in interaction type and object category, which inhibits the modeling of diverse 3D hand-object interaction with the correct physical implication (e.g., contacts and semantics) from text prompts.
To address this challenge, we propose to decompose the interaction generation task into two subtasks: hand-object contact generation; and hand-object motion generation. 
For contact generation, a VAE-based network takes as input a text and an object mesh, and generates the probability of contacts between the surfaces of hands and the object during the interaction. 
The network learns a variety of local geometry structure of diverse objects that is independent of the objects' category, and thus, it is applicable to general objects.
For motion generation, a Transformer-based diffusion model utilizes this 3D contact map as a strong prior for generating physically plausible hand-object motion as a function of text prompts by learning from the augmented labeled dataset; where we annotate text labels from many existing 3D hand and object motion data. 
Finally, we further introduce a hand refiner module that minimizes the distance between the object surface and hand joints to improve the temporal stability of the object-hand contacts and to suppress the penetration artifacts.
In the experiments, we demonstrate that our method can generate more realistic and diverse interactions compared to other baseline methods.
We also show that our method is applicable to unseen objects.
We will release our model and newly labeled data as a strong foundation for future research. 
Codes and data are available in: \href{https://github.com/JunukCha/Text2HOI}{https://github.com/JunukCha/Text2HOI}.

\end{abstract}
\section{Introduction}
\label{Sec:Intro}
Imagine handing over an apple on a table to your friends: you might first grab it and convey this to them. During a social interaction, the hand pose and motion are often defined as a function of object’s pose, shape, and category. While existing works~\cite{petrovich2022temos,guo2022generating,tevet2023human,guo2022tm2t,kim2023flame,zhang2022motiondiffuse,chen2023executing,zhang2023generating} have been successful in modeling diverse and realistic 3D human body motions from a text prompt (where there exists no text-guided hand motion generation works), the context of object interaction has been often missing, which significantly limits the expressiveness in the semantics of the generated motion sequence. In this paper, we propose a first work that can generate realistic and assorted hand-object motions in 3D from a text prompt as shown in Fig.~\ref{fig:teaser}. Our work can be used for various applications such as generating surgical simulations, interactive control of a character for gaming, and future path planning between a robot hand and objects for robotics.

Learning to generate a sequence of 3D hand-object interaction from a text prompt is extremely challenging due to the scarcity of the dataset: the diversity of existing datasets for a sequence of 3D meshes and associated text labels is far behind the one of real-world distribution which is determined by a number of parameters such as hand type (\textit{\eg}, left or right), object's category and structure, scale, contact regions, and so on. A generative model learned from such limited data will fail in the diverse modeling of physically and semantically plausible 3D hand-object interaction.

To overcome this challenge, we propose to decompose the interaction generation task into two subtasks, ``object contact map generation'' and ``hand-object motion generation'', where the models dedicated to each task learn a general geometry representation from the augmented dataset, which leads to the significant improvement in the generalizability and physical plausibility of the combined pipeline.

For contact map generation, we newly develop a contact map prediction network that encodes a local geometry surface of a 3D object mesh along with a target motion text; and generates a 3D contact map---3D probability map at the object's surface that describes the potential regions contacted by hand meshes during the interaction---along with the general geometric features. Since the local geometry representation is category-agnostic, the network is applicable to general objects. By adding condition of the scale information, our contact map generation module is, in nature, able to decode scale-variant probability, \textit{\eg}, if the object's scale is smaller, the region of the predicted contact probability becomes wider, reflecting the natural tendency to grasp smaller objects over a wider area. 

For motion generation, a Transformer-based diffusion model utilizes the contact map and geometric features as strong guidance to generate the sequence of 3D hand and object movements from a text prompt. Unlike a conventional diffusion process~\cite{ho2020denoising}, the model is designed to directly estimate the final sample at each step, which allows us to apply explicit geometric loss (\textit{\eg}, relative distance or orientation) to improve the geometric correctness. Our diffusion model learns the augmented data where we perform extensive manual annotation of the text labels from external motion datasets~\cite{kwon2021h2o, taheri2020grab, fan2023arctic}. 

Using these two modules, we introduce the first text-guided hand-object interaction generation framework that generates the 3D interaction in a compositional way. Given a text prompt, canonical 3D object mesh, and object's scale, our VAE-based contact predictor generates a 3D contact map, and geometry features. Our Transformer-based diffusion model encodes the contact, text, and geometry information with frame-wise and agent-wise (\textit{\ie}, object, and left and right hand) positional embedding to decode realistic 3D hand-object interaction. Finally, our new Transformer-based refiner module pushes the physical correctness of the 3D interaction in a single feed-forward manner by refining the contacts and suppressing the penetration artifacts.

In the experiments, we validate our model on three datasets (H2O~\cite{kwon2021h2o}, GRAB~\cite{taheri2020grab}, and ARCTIC~\cite{fan2023arctic}) where our method outperforms other baseline methods in terms of accuracy, diversity, and physical realism by large margins. We also demonstrate that our compositional framework enables the application of our method to new objects that are not seen during training.

Our contributions can be summarized as follows:
\begin{itemize}
    \item To the best of our knowledge, we propose the first approach that can generate a sequence of 3D hand-object interaction in various styles and lengths from a text prompt. 
    \item We propose a novel compositional framework that enables the modeling of high-quality hand-object interaction from limited data.   
    \item We introduce a new fast and efficient hand refinement module that improves physical realism (\textit{\eg}, penetration-free interaction) without any test-time optimization.
    \item We annotate text labels from existing hands and object motion datasets, which will be made public. 
\end{itemize}

\section{Related Work}
\noindent \textbf{Text to human motion generation.} Thanks to the user-friendly nature of textual inputs, there has been substantial progress in the field of text-guided human motion generation~\cite{petrovich2022temos,guo2022generating,tevet2023human,guo2022tm2t,kim2023flame,zhang2022motiondiffuse,chen2023executing,zhang2023generating,liang2023intergen,lee2023multiact,zhong2023attt2m,jiang2023motiongpt}. Guo~\etal~\cite{guo2022generating} proposed the text2length and text2motion modules to generate human motion in varying time length, while remaining realistic and faithful to the text. Tevet~\etal~\cite{tevet2023human} introduced a Motion Diffusion Model (MDM) for generating natural and expressive human motion, utilizing the geometric losses and Transformer-based approach that predicts the sample instead of noise in each diffusion step. Recently, Liang~\etal~\cite{liang2023intergen} presented a method that can generate interactive motion between two people. But it cannot handle three or more multi-agents.

\begin{figure*}[ht]
\centering
\includegraphics[width=0.99\textwidth]{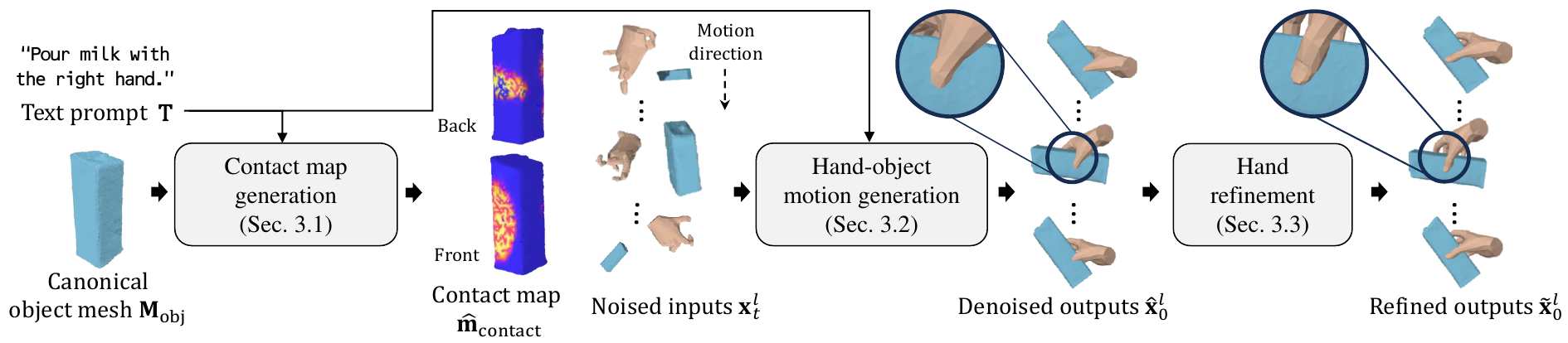}
\caption{\textbf{Schematic diagram of the overall framework.} Given a text prompt and a canonical object mesh prompt, our aim is to generate the 3D motion for hand-object interaction. We first generate a contact map from the canonical object mesh conditioned by the text prompt and object's scale. The hand-object motion generation module removes the noise from the inputs for the denoised outputs to align with the predicted contact map and the text prompt. The denoised outputs exhibit artifacts, including the penetration. To address these artifacts, the hand refinement module adjusts the generated (denoised) hand pose parameters to restrain the penetration and to improve contact interactions.}
\vspace{-3mm}
\label{fig:pipeline}
\end{figure*}

\noindent \textbf{Hand and object motion generation.} 
Existing approaches~\cite{mousavian20196,brahmbhatt2019contactgrasp,corona2020ganhand,karunratanakul2020grasping,jiang2021hand,hampali2020honnotate,grady2021contactopt,zhou2022toch,li2022contact2grasp,brahmbhatt2019contactdb} focus on grasping the stationary object. They are limited in their ability to manipulate the object and are therefore inadequate to generate a natural hand-object motion. Ghosh~\etal~\cite{ghosh2023imos} proposed a method for generating full-body motion in interaction with 3D objects, which is guided by action labels, while it requires an optimization stage for full performance. To generate hand and manipulated object motion, Zhang~\etal~\cite{zhang2021manipnet} proposed a network that relies on the current hand pose, past and future trajectories of both hands and object, and diverse spatial representations. Zheng~\etal~\cite{zheng2023cams} generate the hand-object motion covering both rigid and articulated objects, given an initial hand pose, object geometry, and sparse sequences of object poses. While plausible, these methods~\cite{zhang2021manipnet,zheng2023cams} require the 3D object sequence as inputs, which is often not available from a user. In addition, they cannot utilize text modality.

\begin{figure*}[ht]
\centering
\includegraphics[width=0.99\textwidth]{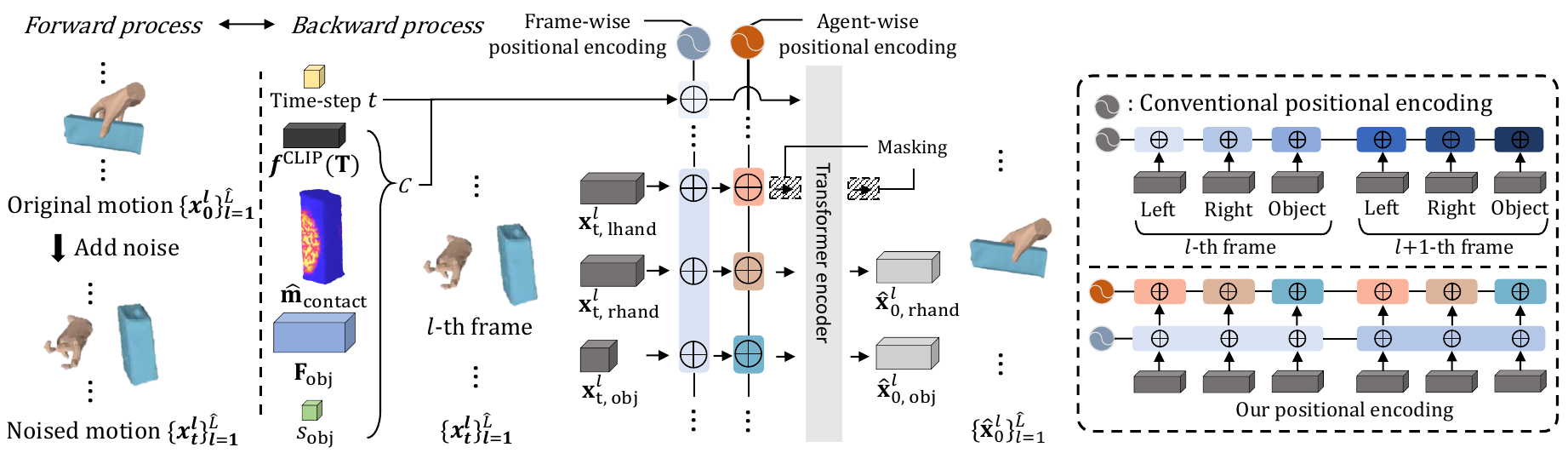}
\vspace{-3mm}
\caption{The details of the text-to-3D hand-object motion generation in our framework. In the forward process, we generate the noised motion $\{\mbx^l_t\}^{\hat{L}}_{l=1}$ by adding the noise to the original (ground-truth) motion $\{\mbx^l_0\}^{\hat{L}}_{l=1}$. In the backward process, the Transformer encoder denoises the noised motion $\{\mbx^l_t\}^{\hat{L}}_{l=1}$, using various conditions $c$ including text features $f^\text{CLIP}(\mbT)$, contact map $\hat{\mbm}_\text{contact}$, object features $\mbF_\text{obj}$, and object's scale $s_\text{obj}$. The right panel illustrates a comparison between conventional positional encoding, which can only differentiate each patch, and our proposed encoding, which provides detailed differentiation of both frames and agents. A unique positional encoding value is assigned for each box, distinguished by different colors.}
\label{fig:pipeline details}
\vspace{-3mm}
\end{figure*}

\section{Method}
Our goal is to generate hand-object interacting motions given a text prompt $\mbT$ and a canonical object mesh $\mbM_\text{obj}$. To address them, we design our framework with three stages, as shown in Fig.~\ref{fig:pipeline}. First, we use the canonical object mesh $\mbM_\text{obj}$ combined with the text feature $f^\text{CLIP}(\mbT)$ via the CLIP text encoder $f^\text{CLIP}$~\cite{radford2021learning} to estimate the contact map $\hat{\mbm}_\text{contact}$ that provides a strong prior for relative 3D locations of hands and an object. Then, we use the Transformer-based diffusion model to denoise the noised input data $\{\mbx^l_t\}_{l=1}^{L}$ at the $t$-th diffusion time-step, where $L$ is the overall sequence length. By conditioning the text features $f^\text{CLIP}(\mbT)$, contact map $\hat{\mbm}_\text{contact}$, object features $\mbF_\text{obj}$ and scale $s_\text{obj}$ on the diffusion model, we estimate the denoised sample $\hat{\mbx}_0$ from the noised one $\mbx_t$. Lastly, hand refiner improves the initial generated hand-object motions considering penetration and contact between hands and an object.

\subsection{Contact map prediction}
\label{sec:method pre-stage}

To generate natural motions for hand-object interaction, it is crucial to understand contact points between hands and an object. For this, we design the contact prediction network $f^\text{contact}$ that encodes contact points on the surfaces of the object mesh $\mbM_\text{obj}$ along with a text prompt $\mbT$ and object's scale $s_\text{obj}$.

We first compute $s_\text{obj}$ which represents the maximum distance from center of object mesh to its vertices. We then sample $N$-point cloud $\mbP \in \mathbb{R}^{N\times3}$ from the vertices of canonical object mesh using the farthest point sampling (FPS) algorithm~\cite{qi2017pointnet}. Subsequently, we normalize $\mbP$ to $\mbP_\text{norm}$ by dividing it with $s_\text{obj}$. The contact prediction network $f^\text{contact}$ receives the normalized point cloud $\mbP_\text{norm}$, text 
features $f^\text{CLIP}(\mbT)$, object's scale $s_\text{obj}$, and Gaussian random noise vector $\mbz_\text{contact}\in\mathbb{R}^{64}$, and produces the contact map $\hat{\mbm}_\text{contact} \in \mathbb{R}^{N\times1}$. In the middle of $f^\text{contact}$, we obtain the object features $\mbF_\text{obj} \in\mathbb{R}^{1,024}$. To train $f^\text{contact}$, we use the combination of binary cross-entropy loss, dice loss and kullback-leibler (KL) divergence loss following~\cite{li2022contact2grasp}.

\subsection{Text-to-3D hand-object motion generation}
\label{sec:method Text2HOI}
Our text-to-3D hand-object interaction generator (Text2HOI) $f^\text{THOI}$, whose architecture is the Transformer encoder~\cite{vaswani2017attention}, is trained via the diffusion-based approach~\cite{ho2020denoising}. 

\subsubsection{Preliminaries.}
The 3D hand-object motion is represented as $\mbx_0=\{\mbx^l_\text{0,lhand}, \mbx^l_\text{0,rhand}, \mbx^l_\text{0,obj}\}_{l=1}^{L_\text{max}}$, where $l$ denotes the frame index. This motion comprises $3\cdot L_\text{max}$ elements, which accounts for the maximum motion length $L_\text{max}$ of three agents (\textit{\ie}, left and right hands and an object): For left and right hands, $\mbx^l_{0, \text{lhand}} \in \mathbb{R}^{99}$ and  $\mbx^l_{0, \text{rhand}} \in \mathbb{R}^{99}$ are composed of $99$-dimensional vectors by flattening and concatenating the 3D hand translation parameters $\mbt^l_{h}\in\mathbb{R}^{3}$ and MANO hand pose parameters $\theta^l\in\mathbb{R}^{16\times6}$ in 6D representation~\cite{zhou2019continuity}. For an object, $\mbx^l_{0, \text{obj}} \in \mathbb{R}^{10}$ is $10$-dimensional vector that concatenates the 3D object translation $\mbt^l_o\in\mathbb{R}^{3}$, object rotation $\mbr^l\in\mathbb{R}^{6}$~\cite{zhou2019continuity}, and object articulation angle $\alpha^l\in\mathbb{R}^{1}$.

The 3D hand-object interaction $\mbx_0$ is used to generate the mesh of hands, and to deform the mesh of objects: The left and right hand meshes are generated from $\mbx_{0, \text{lhand}}$ and $\mbx_{0, \text{rhand}}$ by feeding them to the MANO layer~\cite{romero2017embodied} to output the hand vertices $\mbV_\text{lhand}$, $\mbV_\text{rhand} \in \mathbb{R}^{L\times V \times3}$, and hand joints $\mbJ_\text{lhand}$, $\mbJ_\text{rhand} \in \mathbb{R}^{L\times J \times3}$ in 3D global space, where $V=778$ and $J=21$. A deformed object's point cloud $\mbP_\text{def} \in \mathbb{R}^{L\times N \times3}$ is generated in 3D global space by transforming the object's point clouds $\mbP$ with the translation, rotation and articulation angles in $\mbx_{0, \text{obj}}$. The notation $\hat{\cdot}$ and $\Tilde{\cdot}$ indicate that these values are derived from the estimated $\hat{\mbx}_0$ and refined $\Tilde{\mbx}_0$, respectively.

\subsubsection{Forward process.} Our forward process is formulated as:
\begin{eqnarray}
\mbx_t=\sqrt{\bar{\alpha}_t}\mbx_0+\sqrt{1-\bar{\alpha}_t}\epsilon_t
\end{eqnarray}
following~\cite{ho2020denoising}, where $t$ is the diffusion time-step, $\mbx_0$ is the original 3D hand-object motion, $\mbx_t$ is the noised 3D hand-object motion at the $t$-th diffusion time-step, and $\bar{\alpha}_t \in (0, 1)$ is a set of constant hyper-parameters. The noise $\epsilon_t$ is randomly sampled from the Gaussian distribution at each diffusion-time step $t$.

\subsubsection{Backward process.} In the backward process, the text-to-3D hand-object interaction generator (Text2HOI) $f^\text{THOI}$ denoises the noised motion $\mbx_t$ to reconstruct the original (ground-truth) motion $\mbx_0$: $\hat{\mbx}_0=f^\text{THOI}(\mbx_t, t, c)$, where $c$ denotes the conditions, as described in ~\cite{tevet2023human}. Since we exploit the Transformer encoder as the architecture, the noised signal $\mbx_t$ needs to be first converted to the proper input embedding $\mbX_t$. Similarly, the output of Transformer architecture $\hat{\mbX}_t$ also needs to be converted to the denoised signal $\hat{\mbx}_0$. Furthermore, the text features $f^\text{CLIP}(\mbT)$, object features $\mbF_\text{obj}$, estimated contact map $\hat{\mbm}_\text{contact}$ and object's scale $s_\text{obj}$ are merged together to constitute the conditional signals $\mbX_{\text{cond}}$, which will be detailed in the remainder of the section:

\noindent \textbf{Transformer input generation.} The forwarded signal $\mbx^l_t=\{\mbx^{l}_{t, \text{lhand}}, \mbx^{l}_{t, \text{rhand}}, \mbx^{l}_{t, \text{obj}}\}$ is passed through corresponding fully connected layers (\textit{\ie}, $f^{\text{in}, \text{lhand}}$, $f^{\text{in}, \text{rhand}}$, and $f^{\text{in}, \text{obj}}$), respectively to obtain the input to the Transformer encoder, $\mbX^l_t=\{\mbX^{l}_{t, \text{lhand}} \in \mathbb{R}^{512}, \mbX^{l}_{t, \text{rhand}} \in \mathbb{R}^{512}$, $\mbX^{l}_{t, \text{obj}} \in \mathbb{R}^{512}\}$, respectively. Then, we apply two types of positional encoding: frame-wise and agent-wise. Frame-wise positional encoding adds an sinusoidal value to $\mbX^{l}_{t}$ which varies according to the motion length index $l$; while irrespective to the type of agents. Agent-wise positional encoding adds distinct encoding values for each agent (left hand, right hand, and object), which are consistent across different frames, to $\mbX_{t, \text{lhand}}$, $\mbX_{t, \text{rhand}}$, and $\mbX_{t, \text{obj}}$. These are designed to help the Transformer encoder to better understand the input data. The detail pipeline of these positional encodings is shown in the right bottom panel of Fig.~\ref{fig:pipeline details}.

The Transformer encoder has a maximum input capacity of $451$. The first input is reserved for the conditioning, while the remaining inputs accommodate the maximum motion length $L_\text{max}$ of 150 frames, involving three distinct agents: left hand, right hand and object. We mask out all inputs except for the first $1+3\hat{L}$ inputs where $\hat{L}$ is the estimated length of sequence and subsequently, we mask inputs which are not belonging to the estimated hand type $\mbH^{*}$ (see Sec.~\ref{sec:implementation details} for details about how $\hat{L}$ and $\mbH^{*}$ are estimated).

\noindent \textbf{Conditional input generation.} To generate denoised hand-object motions conditioned on the text prompt $\mbT$ and canonical object mesh $\mbM_\text{obj}$, we need to generate the conditional input for the $t$-th diffusion time-step. Conditional input $\mbX_{t, \text{cond}}$ is generated by:
\begin{eqnarray}
\mbX_{t, \text{cond}} &=& \mbX_\text{cond} + t_\text{emb}
\end{eqnarray}
where the diffusion time-step embedding $t_\text{emb}=f^\text{ts}(t)$ is obtained by applying the diffusion time-step $t$ to the time-step embedding fully-connected layer $f^\text{ts}$ and the condition embedding $\mbX_\text{cond}$ is generated as follows:
\begin{eqnarray}
    \mbX_\text{cond} &=& \mbX^\text{cond}_\text{text}+\mbX^\text{cond}_\text{obj}
\end{eqnarray}
where the text condition $\mbX^\text{cond}_\text{text} =f^\text{text}(f^\text{CLIP}(\mbT))$ is generated by applying the text feature $f^\text{CLIP}(\mbT)$ to the fc layer $f^\text{text}$. The object condition $\mbX^\text{cond}_\text{obj} = f^\text{obj}(\{ \mbF_\text{obj},\hat{\mbm}_\text{contact},s_\text{obj} \})$ is obtained by concatenating object feature $\mbF_\text{obj}$, contact map $\hat{\mbm}_\text{contact}$ and object's scale $s_\text{obj}$, and feeding them to the fc layer $f^\text{obj}$.

\noindent \textbf{Transformer output conversion.} Masked inputs $\mbX_{t}=\{\mbX_{t, \text{cond}}, \mbX^1_{t}, \mbX^2_{t}, \dots, \mbX^{\hat{L}}_{t}\}$ are fed to the Transformer encoder to estimate the outputs ${\hat{\mbX}}_{0}=\{{\hat{\mbX}}^l_{0}\}^{\hat{L}}_{l=1}$, where $\hat{\mbX}^{l}_{0} =\{\hat{\mbX}^{l}_{0, \text{lhand}}, \hat{\mbX}^{l}_{0, \text{rhand}}, \hat{\mbX}^{l}_{0, \text{obj}}\}$. Each outputs—$\hat{\mbX}^{l}_{0, \text{lhand}}$, $\hat{\mbX}^{l}_{0, \text{rhand}}$, and $\hat{\mbX}^{l}_{0, \text{obj}}$— are passed through its own dedicated fully connected layer, denoted as $f^{\text{out}, \text{lhand}}$, $f^{\text{out}, \text{rhand}}$, and $f^{\text{out}, \text{obj}}$, respectively, to obtain the denoised signal $\hat{\mbx}^l_0=\{\hat{\mbx}^{l}_{0, \text{lhand}}, \hat{\mbx}^{l}_{0, \text{rhand}}, \hat{\mbx}^{l}_{0, \text{obj}}\}$.

\noindent \textbf{Training.} Note that the losses related to left and right hands are activated by indicator functions $\mathds{1}_\text{left}$ and $\mathds{1}_\text{right}$, respectively, which are derived from the hand type $\mbH^*$. The $f^\text{THOI}$ is trained with loss functions as follows:
\begin{align}
    L_\text{THOI}(f^\text{THOI}) = L_\text{diff}(f^\text{THOI})+L_\text{dm}(f^\text{THOI})+L_\text{ro}(f^\text{THOI})
\end{align}
where 
\begin{eqnarray}
L_\text{diff}(f^\text{THOI}) = E_{\mbx_t \sim q(\mbx_0|c), t \sim [1, T]} \|\mbx_0-f^\text{THOI}(\mbx_t, t, c)\|^2_2
\end{eqnarray}
is the loss which is used to reconstruct $\mbx_0$ from $\mbx_t$ similar to~\cite{tevet2023human}. We have two more losses (\textit{\ie}, $L_\text{dm}$, $L_\text{ro}$) to make the $f^\text{THOI}$ to generate more accurate hand-object motions. The distance map loss $L_\text{dm}$, proposed in~\cite{liang2023intergen}, is employed in our hand-object motion generation to align the estimated distance map with ground-truth distance map as follows:
\begin{align}
L_\text{dm}(f^\text{THOI}) = \sum_{i=1}^{\hat{L}\times J\times N} \bigg\{&\mathds{1}_\text{left} \cdot \bigg((\hat{\mbd}^i_\text{left}-\mbd^i_\text{left})\cdot I(\mbd^i_\text{left}<\tau)\bigg)^2 \nonumber\\
+ &\mathds{1}_\text{right} \cdot \bigg((\hat{\mbd}^i_\text{right}-\mbd^i_\text{right})\cdot I(\mbd^i_\text{right}<\tau)\bigg)^2\bigg\}
\end{align}
where $\hat{\mbd}^i_\text{left}$ and $\hat{\mbd}^i_\text{right}$ denote the $i$-th element of $\hat{\mbd}_\text{left}$ and $\hat{\mbd}_\text{right} \in \mathbb{R}^{\hat{L}\times J\times N}$, respectively. These represent the estimated distance maps between the $J$ hand joints (left $\hat{\mbJ}_\text{lhand}$ and right $\hat{\mbJ}_\text{rhand}$) and the $N$ object points $\hat{\mbP}_\text{def}$ across a sequence of $\hat{L}$ frames, derived from their 3D global positions. $\mbd^i_\text{left}$ and $\mbd^i_\text{right}$ denote the $i$-th element of $\mbd_\text{left}$ and $\mbd_\text{right} \in \mathbb{R}^{\hat{L}\times J\times N}$ which are the ground-truth distance maps obtained for left and right hands, respectively. The indicator function $I(\cdot)$ outputs $1$ when the statement is true and $0$, otherwise. It activates the loss only when the hand-object distance is below the distance threshold $\tau$, where it is empirically set as $2cm$.

In the relative orientation loss $L_\text{ro}$, we consider the 3D relative rotation as follows, as hands and objects exhibit severe rotation changes:
\begin{align}
    L_\text{ro}(f^\text{THOI}) = &\mathds{1}_\text{left} \cdot \|R(\hat{\mbx}_{0, \text{lhand}}, \hat{\mbx}_{0, \text{obj}})-R(\mbx_{0, \text{lhand}}, \mbx_{0, \text{obj}})\|^2_2 \ \nonumber \\
    + &\mathds{1}_\text{right} \cdot \|R(\hat{\mbx}_{0, \text{rhand}}, \hat{\mbx}_{0, \text{obj}})-R(\mbx_{0, \text{rhand}}, \mbx_{0, \text{obj}})\|^2_2,
\end{align}
where $R(\cdot,\cdot)$ indicates the 3D relative orientation between hand and object.  

\noindent \textbf{Sampling.} At each time-step $t$, the model $f^\text{THOI}$ predicts a clean motion, denoted as $\hat{\mbx}_0 = f^\text{THOI}(\mbx_t, t, c)$, and then re-noise $\hat{\mbx}_0$ to $\mbx_{t-1}$ \cite{tevet2023human}. This procedure is conducted repeatedly, starting from $t=T$ to $t=1$.

\subsection{Hand refinement network}
\label{sec: method hand refinement}
We propose a hand refinement network $f^\text{ref}$ that considers the contact and penetration between hands and an object generated from Text2HOI $f^\text{THOI}$ in Sec.~\ref{sec:method Text2HOI}. The architecture of $f^\text{ref}$ is similar to that of $f^\text{THOI}$: 1) it employs a Transformer encoder architecture, and 2) it utilizes frame-wise and agent-wise position encoding. The main differences between $f^\text{THOI}$ and $f^\text{ref}$ are that $f^\text{ref}$ does not involve the diffusion mechanism; it does not receive any conditions as input; and it refines only hand motions.

\noindent \textbf{Inputs and outputs.} The hand refinement network receives several inputs: Text2HOI's hand output $\hat{\mbx}_{0, \text{hand}}$, hand joints $\hat{\mbJ}_\text{hand}$, predicted contact map $\hat{\mbm}_\text{contact}$, deformed object's point cloud $\hat{\mbP}_\text{def}$, and distance-based attention map $\mbm_\text{att}$. The attention map $\mbm_\text{att} = \exp(- 50 \times \mbD)$ is defined as~\cite{taheri2023grip}, where $\mbD \in \mathbb{R}^{J\times3}$ represents the 3D displacement between $J$ hand joints $\hat{\mbJ}_\text{hand}$ and the nearest object points in $\hat{\mbP}_\text{def}$. These components, denoted as $\hat{\mbx}_{0, \text{hand}}$, $\hat{\mbJ}_\text{hand}$, $\hat{\mbm}_\text{contact}$, $\hat{\mbP}_\text{def}$, and $\mbm_\text{att}$, are flattened and concatenated to form the hand refiner input. As indicated in Sec.~\ref{sec:method Text2HOI}, these inputs are masked using $\mbH^{*}$. Then, $f^\text{ref}$ outputs the refined hand motions $\Tilde{\mbx}_\text{hand}$. They are masked using $\mbH^*$ for loss calculation and result visualization.

\begin{table*}[t]
    \centering
    \caption{Comparison on H2O, GRAB, and ARCTIC datasets. $\dag$ denotes our produced results. $\rightarrow$ denotes that the higher value of the metric, the closer to the GT distribution. Best results are emphasized in bold.}
    \vspace{-3mm}
    \begin{tabular}{lccccc}
    \hline
    & \multicolumn{5}{c}{H2O} \\ 
    Method & Accuracy (top-3) $\uparrow$ & FID $\downarrow$ & Diversity $\rightarrow$ & Multimodality $\uparrow$ & Physical realism $\uparrow$ \\
    \hline
    GT & 0.9920 $\pm$ 0.0003 & - & 0.6057 $\pm$ 0.0050 & 0.2067 $\pm$ 0.0024 & 0.4790 $\pm$ 0.0002 \\
    \hline
    $\text{T2M}^\dag$~\cite{guo2022generating} & 0.6463 $\pm$ 0.0014 & 0.3439 $\pm$ 0.0006 & 0.3475 $\pm$ 0.0040 & 0.0634 $\pm$ 0.0022 & 0.3890 $\pm$ 0.016 \\
    $\text{MDM}^\dag$~\cite{tevet2023human} & 0.5832 $\pm$ 0.0011 & 0.3015 $\pm$ 0.0011 & 0.5127 $\pm$ 0.0054 & 0.1738 $\pm$ 0.0049 & 0.5572 $\pm$ 0.0013 \\
    $\text{IMOS}^\dag$~\cite{ghosh2023imos} & 0.5518 $\pm$ 0.0026 & 0.2945 $\pm$ 0.0011 & 0.4076 $\pm$ 0.0056 & 0.1798 $\pm$ 0.0115 & 0.3532 $\pm$ 0.0026 \\
    Ours & \textbf{0.8295 $\pm$ 0.0015} & \textbf{0.1744 $\pm$ 0.0013} & \textbf{0.5365 $\pm$ 0.0073} & \textbf{0.2469 $\pm$ 0.0081} & \textbf{0.7574 $\pm$ 0.0022} \\
    \hline
    \hline
    & \multicolumn{5}{c}{GRAB}\\ 
    \hline
    GT & 0.9994 $\pm$ 0.0001 & - & 0.8557 $\pm$ 0.0054 & 0.4390 $\pm$ 0.0045 & 0.8084 $\pm$ 0.0002 \\
    \hline
    $\text{T2M}^\dag$~\cite{guo2022generating} & 0.1897 $\pm$ 0.0007 & 0.7886 $\pm$ 0.0005 & 0.5712 $\pm$ 0.0078 & 0.0964 $\pm$ 0.0027 & 0.5844 $\pm$ 0.0002 \\
    $\text{MDM}^\dag$~\cite{tevet2023human} & 0.5127 $\pm$ 0.0009 & 0.6023 $\pm$ 0.0011 & 0.8012 $\pm$ 0.0054 & 0.5194 $\pm$ 0.0145 & 0.7382 $\pm$ 0.0004 \\
    $\text{IMOS}^\dag$~\cite{ghosh2023imos} & 0.4097 $\pm$ 0.0005 & 0.6147 $\pm$ 0.0003 & 0.6861 $\pm$ 0.0060 & 0.2845 $\pm$ 0.0036 & 0.6418 $\pm$ 0.0014 \\
    Ours & \textbf{0.9218 $\pm$ 0.0010} & \textbf{0.3017 $\pm$ 0.0004} & \textbf{0.8351 $\pm$ 0.0061} & \textbf{0.5216 $\pm$ 0.0131} & \textbf{0.8839 $\pm$ 0.0005} \\
    \hline
    \hline
    & \multicolumn{5}{c}{ARCTIC}\\ 
    \hline
    GT & 0.9997 $\pm$ 0.0001 & - & 0.5916 $\pm$ 0.0037 & 0.3279 $\pm$ 0.0038 & 0.9573 $\pm$ 0.0000 \\
    \hline
    $\text{T2M}^\dag$~\cite{guo2022generating} & 0.5234 $\pm$ 0.0015 & 0.3599 $\pm$ 0.0005 & 0.3301 $\pm$ 0.0023 & 0.0849 $\pm$ 0.0017 & 0.0143 $\pm$ 0.0001 \\
    $\text{MDM}^\dag$~\cite{tevet2023human} & 0.5572 $\pm$ 0.0012 & 0.3025 $\pm$ 0.0006 & 0.4984 $\pm$ 0.0039 & 0.2632 $\pm$ 0.0065 & 0.7043 $\pm$ 0.0009 \\
    $\text{IMOS}^\dag$~\cite{ghosh2023imos} & 0.8190 $\pm$ 0.0039 & 0.1826 $\pm$ 0.0005 & 0.5702 $\pm$ 0.0039 & 0.2741 $\pm$ 0.0049 & 0.7569 $\pm$ 0.0023 \\
    Ours & \textbf{0.9205 $\pm$ 0.0012} & \textbf{0.1329 $\pm$ 0.0006} & \textbf{0.5758 $\pm$ 0.0042} & \textbf{0.3170 $\pm$ 0.0068} & \textbf{0.8760 $\pm$ 0.0009} \\
    \hline
    \end{tabular}
    \label{tab:comparison_sota}
    \vspace{-3mm}
\end{table*}

\noindent \textbf{Training.} The hand refinement network is trained using the loss function $L_\text{refine}$ as follows:
\begin{align}
    L_\text{refine}(f^\text{ref})=L_\text{simple}(f^\text{ref})+ L_\text{penet}(f^\text{ref})+\lambda_1 L_\text{contact}(f^\text{ref}),
\end{align}
where $\lambda_1$ is set as 5. The simple L2 loss is expressed as follows:
\vspace{-3mm}
\begin{align}
    L_\text{simple} (f^\text{ref}) = \| \Tilde{\mbx}_\text{hand} - \mbx_\text{hand} \|^2_2,
\end{align}
where $\mbx_\text{hand}$ denotes the ground-truth hand motions. The penetration loss $L_\text{penet}$~\cite{jiang2021hand} is applied only on hand vertices that penetrate the object surfaces as follows:
\begin{align}
    L_\text{penet}(f^\text{ref}) = \mathds{1}_\text{left} \cdot ||d(\Tilde{v}_\text{lhand}, \hat{p}^\text{left}_\text{obj})||^2 + \mathds{1}_\text{right} \cdot ||d(\Tilde{v}_\text{rhand}, \hat{p}^\text{right}_\text{obj})||^2,
\end{align}
where $d(\cdot, \cdot)$ denotes the Euclidean distance between two points, $\Tilde{v}_\text{lhand}\in\Tilde{\mbV}_\text{lhand}$ and $\Tilde{v}_\text{rhand}\in\Tilde{\mbV}_\text{rhand}$ are hand vertices that penetrate the object surface, and $\hat{p}^\text{left}_\text{obj}\in\hat{P}_\text{def}$ and $\hat{p}^\text{right}_\text{obj}\in\hat{P}_\text{def}$ denote the object points closest to  $\Tilde{v}_\text{lhand}$ and $\Tilde{v}_\text{rhand}$, respectively. The contact loss $L_\text{contact}$~\cite{jiang2021hand} is applied to the joints that are sufficiently close to the object surface, as follows:
\begin{align}
    L_\text{contact}(f^\text{ref}) = \mathds{1}_\text{left} \cdot ||d(\Tilde{j}_\text{lhand}, \hat{c}^\text{left}_\text{obj})||^2 + \mathds{1}_\text{right } \cdot ||d(\Tilde{j}_\text{rhand}, \hat{c}^\text{right}_\text{obj})||^2,
\end{align}
where $\Tilde{j}_\text{lhand}\in\Tilde{\mbJ}_\text{lhand}$ and $\Tilde{j}_\text{rhand}\in\Tilde{\mbJ}_\text{rhand}$ represent the hand joints that are within a distance threshold $\tau$ from the object surface, respectively. $\hat{c}^\text{left}_\text{obj}\in\hat{P}_\text{def}$ and $\hat{c}^\text{right}_\text{obj}\in\hat{P}_\text{def}$ represent object points closest to  $\Tilde{j}_\text{lhand}$ and $\Tilde{j}_\text{rhand}$, respectively.

\section{Experiments}

\subsection{Implementation details}
\label{sec:implementation details}
We use $T$ = 1,000 noising steps and a cosine noise schedule. We use sinusoidal positional encoding for frame-wise and agent-wise positional encodings. We set the maximum length of motion sequences, denoted as $L_\text{max}$, to 150 frames. Further details about network architecture can be found in the supplemental material.

\noindent \textbf{Hand-type selection.} We use the CLIP text encoder~\cite{radford2021learning} $f^\text{CLIP}$ to calculate cosine similarity between the input text prompt $\mbT$ and predefined prompt templates $\Gamma(\mbH)$=``A photo of $\mbH$'', where $\mbH \in \{\text{left hand}, \text{right hand}, \text{both hands}\}$. The hand type $\mbH^{*}$ with the highest cosine similarity to $\mbT$ is selected for masking in the Transformer's inputs, outputs, and losses (see Secs.~\ref{sec:method Text2HOI}, \ref{sec: method hand refinement} and supplemental material.).


\noindent \textbf{Motion length prediction.} To obtain proper motion length $\hat{L}\le L_\text{max}$, we design a motion-length prediction network $f^\text{Length}$. It receives the text feature vector $f^\text{CLIP}(\mbT)$ and Gaussian random noise $\mbn\in\mathbb{R}^{64}$, to predict the appropriate motion length $\hat{L}$ for the text prompt $\mbT$. To train $f^\text{Length}$, we use the loss function $L_\text{length} = \|\hat{L}-L\|^2$, where $L$ is ground truth.

\begin{table*}[ht]
    \centering
    \caption{Ablation study on the positional encoding, losses, and conditions for `Ours w/o $f^\text{ref}$' and ablation study on losses for `Ours'.}
    \vspace{-3mm}
    \resizebox{1\textwidth}{!}{
    \begin{tabular}{lcccccc}
    \hline
    & & \multicolumn{5}{c}{GRAB}\\ 
    Method & $f^\text{ref}$ & Accuracy (top-3) $\uparrow$ & FID $\downarrow$ & Diversity $\rightarrow$ & Multimodality $\uparrow$ & Physical realism $\uparrow$ \\
    \hline
    GT & - & 0.9994 $\pm$ 0.0001 & - & 0.8557 $\pm$ 0.0054 & 0.4390 $\pm$ 0.0045 & 0.8084 $\pm$ 0.0002 \\
    \hline
    w/o frame-wise \& agent-wise PE  & \xmark & 0.8294 $\pm$ 0.0016 & 0.3461 $\pm$ 0.0018 & 0.7814 $\pm$ 0.063 & 0.4776 $\pm$ 0.0194 & 0.8024 $\pm$ 0.0007 \\
    w/o agent-wise PE & \xmark & 0.8314 $\pm$ 0.0012 & 0.3412 $\pm$ 0.0006 & 0.8011 $\pm$ 0.067 & 0.4755 $\pm$ 0.0122 & 0.8221 $\pm$ 0.0009 \\
    \hline
    w/o $L_\text{dm}$ \& $L_\text{ro}$ & \xmark & 0.8289 $\pm$ 0.0038 & 0.3416 $\pm$ 0.0020 & 0.7887 $\pm$ 0.0640 & 0.4654 $\pm$ 0.0150 & 0.7490 $\pm$ 0.0006 \\
    w/o $L_\text{ro}$ & \xmark & 0.8272 $\pm$ 0.0020 & 0.3407 $\pm$ 0.0015 & 0.7997 $\pm$ 0.0079 & 0.4627 $\pm$ 0.0104 & 0.8247 $\pm$ 0.0011 \\
    w/o $L_\text{dm}$ & \xmark & 0.8202 $\pm$ 0.0017 & 0.3444 $\pm$ 0.0007 & \textbf{0.8156 $\pm$ 0.0070} & 0.4819 $\pm$ 0.0125 & 0.6410 $\pm$ 0.0010 \\
    \hline
    w/o $\hat{\mbm}_\text{contact}$ \& $s_\text{obj}$ & \xmark & 0.8197 $\pm$ 0.0009 & 0.3428 $\pm$ 0.0012 & 0.7994 $\pm$ 0.0055 & 0.4305 $\pm$ 0.0121 & 0.7815 $\pm$ 0.0006 \\
    w/o $s_\text{obj}$ & \xmark & 0.8274 $\pm$ 0.0013 & 0.3413 $\pm$ 0.0010 & 0.7963 $\pm$ 0.0054 & 0.4405 $\pm$ 0.0139 & 0.8018 $\pm$ 0.0005 \\
    w/o $\hat{\mbm}_\text{contact}$ & \xmark & 0.8277 $\pm$ 0.0027 & 0.3411 $\pm$ 0.0006 & 0.8012 $\pm$ 0.0067 & 0.4455 $\pm$ 0.0115 & 0.7892 $\pm$ 0.0009 \\
    \hline
    Ours w/o $f^\text{ref}$ & \xmark & \textbf{0.8411 $\pm$ 0.0009} & \textbf{0.3321 $\pm$ 0.0006} & 0.8143 $\pm$ 0.0050 & \textbf{0.4989 $\pm$ 0.0154} & \textbf{0.8312 $\pm$ 0.0005} \\
    \hline
    \hline
    w/o $L_\text{penet}$ \& $L_\text{contact}$ & \cmark & 0.8838 $\pm$ 0.0014 & 0.3234 $\pm$ 0.0007 & 0.8277 $\pm$ 0.0068 & 0.5111 $\pm$ 0.014 & 0.6249 $\pm$ 0.0008 \\
    w/o $L_\text{contact}$ & \cmark & 0.8827 $\pm$ 0.0008 & 0.3114 $\pm$ 0.0013 & 0.8301 $\pm$ 0.0048 & 0.4808 $\pm$ 0.0151 & 0.1467 $\pm$ 0.0005 \\
    w/o $L_\text{penet}$ & \cmark & 0.8941 $\pm$ 0.0009 & 0.3024 $\pm$ 0.0005 & 0.8267 $\pm$ 0.0061 & 0.5182 $\pm$ 0.0099 & 0.8782 $\pm$ 0.0006 \\
    \hline
    Ours & \cmark & \textbf{0.9218 $\pm$ 0.0010} & \textbf{0.3017 $\pm$ 0.0004} & \textbf{0.8351 $\pm$ 0.0061} & \textbf{0.5216 $\pm$ 0.0131} & \textbf{0.8839 $\pm$ 0.0005} \\
    \hline
    \end{tabular}}
    \label{tab:ablation}
    \vspace{-3mm}
\end{table*}

\subsection{Dataset}
We use H2O~\cite{kwon2021h2o}, GRAB~\cite{taheri2020grab}, and ARCTIC~\cite{fan2023arctic} in our experiment, which collects hand-object mesh sequences. We automatically generate text prompts by exploiting action labels for H2O and GRAB datasets; while we manually label text prompts for ARCTIC. The characteristics of three datasets and details of our annotation process are further illustrated in the supplemental material.

\subsection{Evaluation metrics and baselines}
\noindent \textbf{Evaluation metrics.} We use the metrics of accuracy, frechet inception distance (FID), diversity, and multi-modality, as used in IMOS~\cite{ghosh2023imos}. The accuracy serves as an indicator of how well the model generates motions and is evaluated by the pre-trained action classifier. We train a standard RNN-based action classifier to extract motion features and classify the action from the motions, as in IMOS~\cite{ghosh2023imos}. The FID quantifies feature-space distances between real and generated motions, capturing the dissimilarity. The diversity reflects the range of distinct motions, and multi-modality measures the average variance of motions for an individual text prompt. To assess the physical realism of generated hand-object motions, we employ a physical model following the approach in ManipNet~\cite{zhang2021manipnet}, assigning a realism score of $0$ (unreal) or $1$ (real) for measuring the realism of each frame. Experiments are conducted $20$ times to establish the robustness, and we reported results within a 95\% confidence interval.

\noindent \textbf{Baselines.} We compare our approach with three existing text-to-human motion generation methods: T2M~\cite{guo2022generating}, MDM~\cite{tevet2023human}, and IMOS~\cite{ghosh2023imos}. T2M~\cite{guo2022generating} employs a temporal VAE-based architecture and MDM~\cite{tevet2023human} utilizes a diffusion model. IMOS~\cite{ghosh2023imos} is designed to first generate human body and arm motions conditioned on both action labels and past body motions. It then optimizes object rotation and translation based on their history to generate body and arm motion. Since they were originally designed for generating individual human motions from text prompts, to ensure a fair comparison, we re-train the methods using hand-object motion data, allowing them to generate hand-object motions from text prompts.

\begin{figure}[t]
\centering
\includegraphics[width=0.47\textwidth]{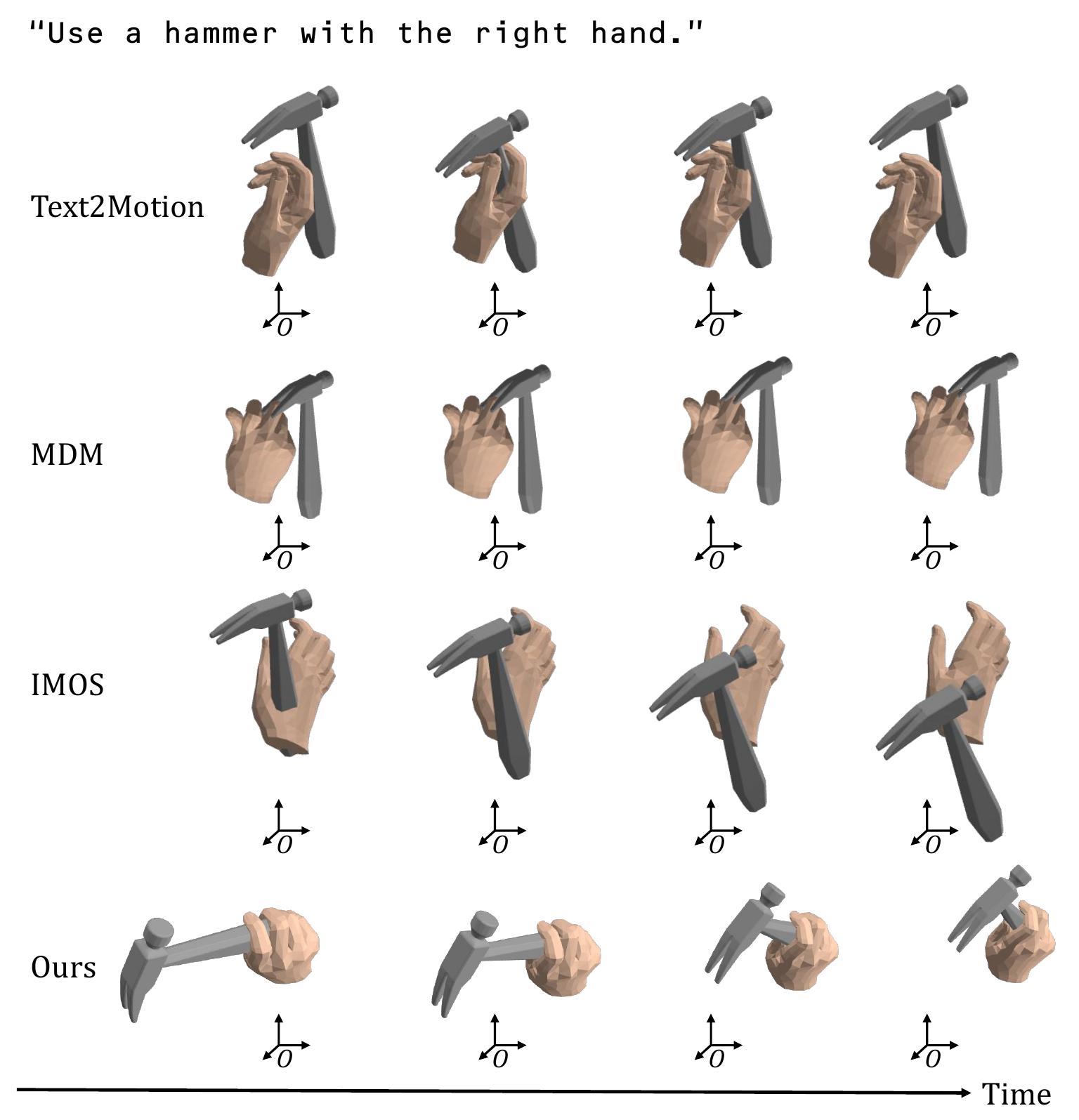}
\vspace{-4mm}
\caption{We compare our generated hand-object motions with other baselines' results. Each row show the results of Text2Motion~\cite{guo2022generating}, MDM~\cite{tevet2023human}, IMOS~\cite{ghosh2023imos}, and ours.}
\label{fig:qualitative comparsion}
\vspace{-5mm}
\end{figure}

\subsection{Experimental results}
\noindent \textbf{Comparison to other methods.} We compare our method with other state-of-the-art methods (\textit{\ie},  T2M~\cite{guo2022generating}, MDM~\cite{tevet2023human}, and IMOS~\cite{ghosh2023imos}), as shown in Tab.~\ref{tab:comparison_sota}. For all datasets, our method outperforms other baselines in multiple measures. Particularly, our method demonstrates exceptional performance in generating physically realistic hand-object motions, as evidenced by achieving the highest score in Physical realism compared to other approaches. The similarity of our distribution to the ground truth (GT) distribution in terms of Diversity, along with our highest scores in Multimodality and Accuracy, demonstrates our model's capability to generate motions that are both diverse and accurate, and are well-aligned with text prompts.
We compare our qualitative results with other baselines in Fig.~\ref{fig:qualitative comparsion}. It shows that our method, compared to others, outperforms in generating motions where hand and object interact realistically, and these motions align closely with text prompts \textit{``Use a hammer with the right hand."}. The right hand well grabs the hammer and mimics the motion of driving something into a wall.

\begin{figure*}[t]
\centering
\includegraphics[width=0.97\textwidth]{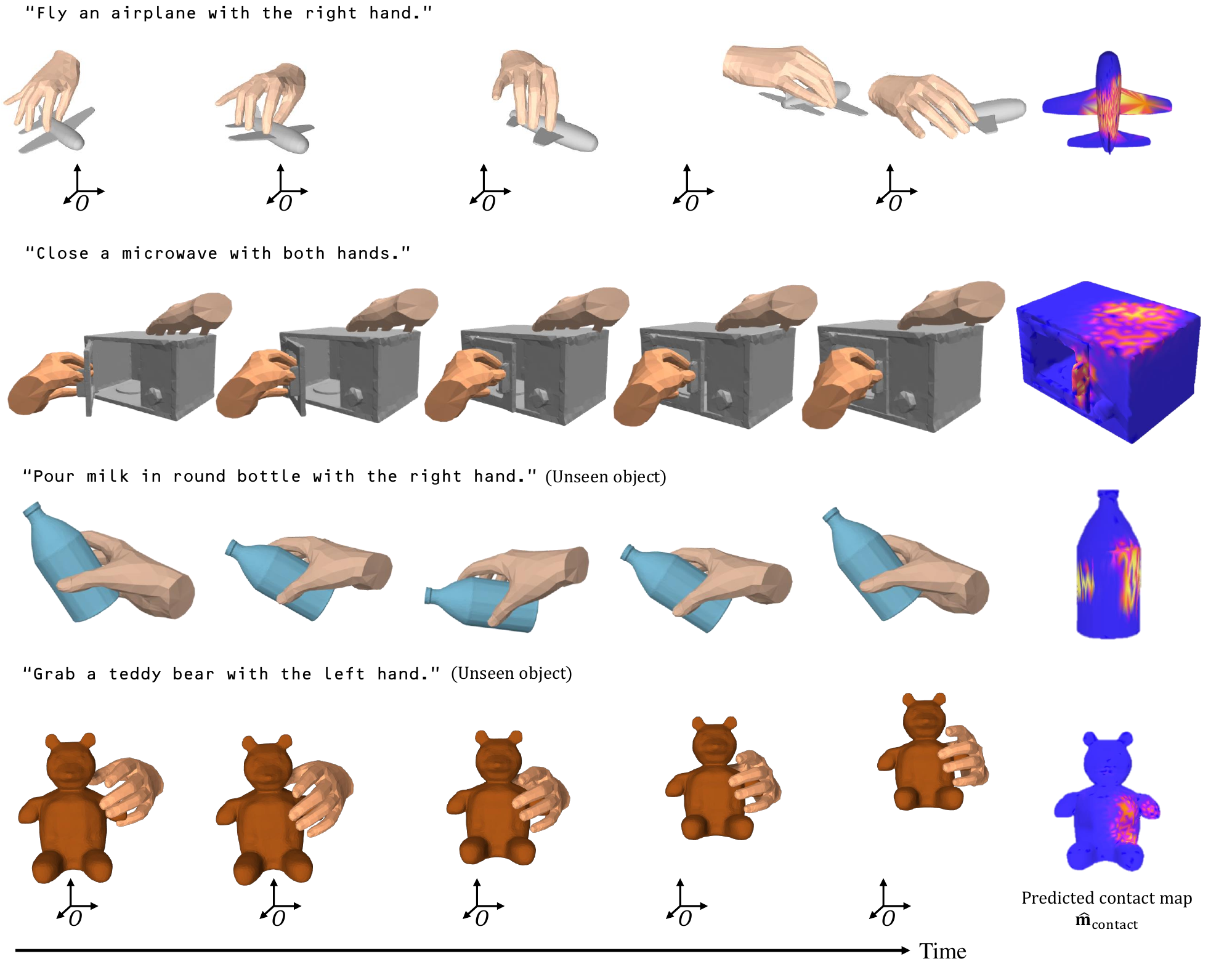}
\vspace{-3mm}
\caption{We demonstrate the generated hand-object motions and the predicted contact map results. The first and second rows show the results with objects seen during training. The third and fourth rows show the results with objects unseen during training.} 
\vspace{-5mm}
\label{fig:qualitative}
\end{figure*}

\noindent \textbf{Qualitative results} are shown in Fig.~\ref{fig:qualitative}. Our method generates realistic hand-object motions that are closely aligned with the input text prompts, effectively handling even unseen objects. Please refer to the supplemental material for more visualizations including video results, and text-guided and scale-variant contact maps.

\vspace{-1mm}
\subsection{Ablation study}
We conduct several ablation studies on GRAB dataset, to validate the effectiveness of our modules. The results are demonstrated in Tab.~\ref{tab:ablation}. 

\noindent\textbf{Position encoding.} We introduce two types of positional encodings: frame-wise and agent-wise, which assists the Transformer to interpret inputs in a more distinct way. Seeing the results `w/o frame-wise \& agent-wise PE' and `w/o agent-wise PE', we can conclude that by leveraging the specialized positional encoding, $f^\text{THOI}$ is capable of generating more realistic hand-object motions. 

\noindent\textbf{Losses.} We remove the distance map loss $L_\text{dm}$ and relative orientation loss $L_\text{ro}$ in our approach, and see how the performance changes: Seeing `w/o $L_\text{dm}$ \& $L_\text{ro}$', `w/o $L_\text{dm}$' and `w/o $L_\text{ro}$', we can conclude that these losses induce better results by facilitating model's understanding of 3D relationship between hands and an object.

\noindent \textbf{Condition inputs.} We remove the contact map $\hat{\mbm}_\text{contact}$ and scale of the object $s_\text{obj}$ conditions from the original pipeline, and see how the performance changes. Seeing `w/o $\hat{\mbm}_\text{contact} \& s_\text{obj}$', `w/o $s_\text{obj}$' and `w/o $\hat{\mbm}_\text{contact}$', we can observe that gradually including additional conditions aids in generating more appropriate hand poses to the object.

\noindent\textbf{Refiner.} Compared to the `Ours w/o $f^\text{refiner}$' that does not involve the refiner, `Ours' provides far better performance, especially in the physical realism of hand and object motions. Also, we demonstrate the effect of losses $L_\text{penet}$ and $L_\text{contact}$ by removing them in `w/o $L_\text{penet}$ \& $L_\text{contact}$', `w/o $L_\text{contact}$' and `w/o $L_\text{penet}$', as shown in the same table. Involving more losses consistently improve the performance. 

\section{Conclusion}
In this paper, we propose a novel method for generating the sequence of 3D hand-object interaction from a text prompt and a canonical object mesh. This is achieved through the three-staged framework that 1) estimates the text-guided and scale-variant contact maps; 2) generates hand-object motions based on a Transformer-based diffusion mechanism; and 3) refines the interaction by considering the penetration and contacts between hands and an object. In experiments, we validate our effectiveness of hand-object interaction generation by comparing it to three baselines 
where our method outperforms previous methods with strong physical plausibility and accuracy. 


\noindent \textbf{Acknowledgements.} This work was supported by IITP grants (No. 2020-0-01336 Artificial intelligence graduate school program (UNIST) 10\%; No. 2021-0-02068 AI innovation hub 10\%; No. 2022-0-00264 Comprehensive video understanding and generation with knowledge-based deep logic neural network 20\%) and the NRF grant (No. RS-2023-00252630 20\%), all funded by the Korean government (MSIT). This work was also supported by Korea Institute of Marine Science \& Technology Promotion (KIMST) funded by Ministry of Oceans and Fisheries (RS-2022-KS221674) 20\% and received support from AI Center, CJ Corporation (20\%).

{
    \small
    \bibliographystyle{ieeenat_fullname}
    \bibliography{main}
}


\twocolumn[{
\begin{center}
\textbf{\Large Text2HOI: Text-guided 3D Motion Generation for Hand-Object Interaction\\-Supplementary-\\}
\vspace{10mm}
\Large \author{Junuk Cha\textsuperscript{1} \qquad Jihyeon Kim\textsuperscript{1,2\dag} \qquad Jae Shin Yoon\textsuperscript{3*} \qquad Seungryul Baek\textsuperscript{1*} \vspace{0.3em} \\
{\normalsize $^1$UNIST} \qquad
{\normalsize $^2$KETI} \qquad
{\normalsize $^3$Adobe Research}
}
\end{center}
\vspace{16mm}
}]

\setcounter{equation}{0}
\setcounter{figure}{0}
\setcounter{table}{0}
\setcounter{page}{1}
\makeatletter
\renewcommand{\theequation}{S\arabic{equation}}
\renewcommand{\thefigure}{S\arabic{figure}}
\renewcommand{\thetable}{S\arabic{table}}

\blfootnote{This research was conducted when Jihyeon Kim was a graduate student (Master candidate) at UNIST$\dag$. Co-last authors$*$.}

In this supplementary material, we provide detailed descriptions of text annotation process along with data summary; qualitative results for ablation studies; refiner design; the effect of the refiner; inference speed; the summary of notations; network architectures utilized in our pipeline; masking process; implementation for baselines; articulation angle; and limitation. Additionally, for further results (the qualitative comparison and the additional video results), please refer to the accompanying supplementary video. The video is also available in: \href{https://youtu.be/YBRsu0pnTeA}{https://youtu.be/YBRsu0pnTeA}.

\section{Dataset}
\noindent \textbf{Text prompt annotation.} The datasets H2O~\cite{kwon2021h2o}, GRAB~\cite{taheri2020grab}, and ARCTIC~\cite{fan2023arctic} do not provide the text prompts which describe the hand-object interactions. To facilitate the generation of 3D hand-object interactions from text prompts, collecting such text prompts is necessary. Text prompts should include details about the interacting hand type (\textit{\eg}, left hand, right hand, and both), the action involved (\textit{\eg}, grab, place, lift), and the object category or name (\textit{\eg}, apple, airplane, microwave). The basic format for text prompts is as follows: ``\{action\}~ \{object category\} with~ \{hand type\}." (\textit{\eg}, ``Place a book with both hands."). For the H2O and GRAB datasets, we automatically annotate text using provided action labels, which include both the action and object category. However, these labels do not specify the interacting hand type. We determine the interacting hand type based on the proximity between the hand and the object in global 3D space; if the distance during interaction is less than a predefined threshold (2cm), we decide that the hand is involved in the interaction motion. For the ARCTIC dataset, we conduct manual text annotation by observing the video, noting the action, hand type, and object category. We further augment the text prompt with additional descriptors using passive form, subject modifications, and gerunds. For instance, prompts are augmented to formats like ``A~ \{object category\}~is~\{action\}~by~\{hand type\}." (\textit{\eg} ``A book is placed by both hands."), or ``\{hand type\}~ \{action\}~ \{object category\}." (\textit{\eg} ``Both hands place a book.") or ``\{action\}$+$ing~ \{object category\}~with~ \{hand type\}." (\textit{\eg} ``Placing a book with both hands."). 

\noindent \textbf{H2O.} The H2O dataset~\cite{kwon2021h2o} consists of 660 interactions, each involving two hands and one of 8 distinct objects. The dataset is annotated with 11 unique verb labels. Using these object and verb labels, and identifying which hand is interacting with the object, we automatically generate 272 distinct sentences to represent these interactions. Additionally, the dataset provides MANO hand parameters, object meshes, as well as information on object rotation and translation.

\noindent \textbf{GRAB.} The GRAB dataset~\cite{taheri2020grab} consists of 1,335 motions involving two hands and one of 51 distinct objects of varying shape and size. The dataset has 29 action labels. Using these object and action labels, and identifying which hand is interacting with the object, we automatically generate 1,104 distinct sentences to represent these motions. Additionally, the dataset provides MANO hand parameters, object meshes, as well as information on object rotation and translation.

\noindent \textbf{ARCTIC.} The ARCTIC dataset~\cite{fan2023arctic} is released for reconstructing hands and objects from RGB images. We annotate the data with our defined 11 action labels. We manually create 644 sentences to describe the motions, and these sentences contain information about the action, the type of object, and which hand is interacting. We annotate a total of 4,597 motions. Essentially, this dataset provides MANO hand parameters, object meshes, as well as information on object rotation, translation, and articulation angle with the pre-defined axis.

\begin{figure}[t!]
\centering
\includegraphics[width=0.46\textwidth]{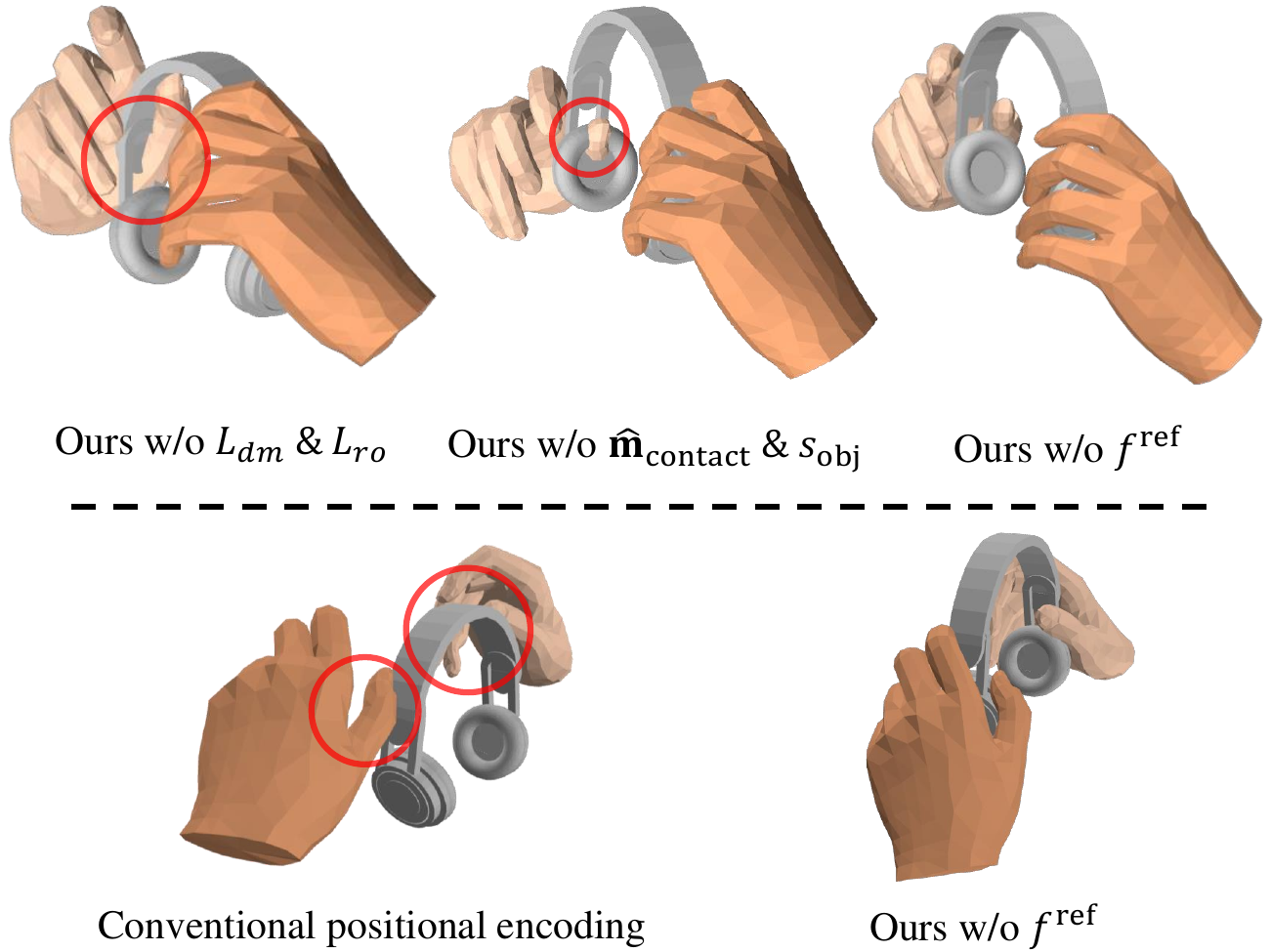}
\caption{In the first row, the comparisons of geometry losses and conditions are presented. In the second row, the comparison focuses on positional encodings.}
\label{fig:ablation THOI}
\end{figure}

\begin{figure}[t!]
\centering
\includegraphics[width=0.46\textwidth]{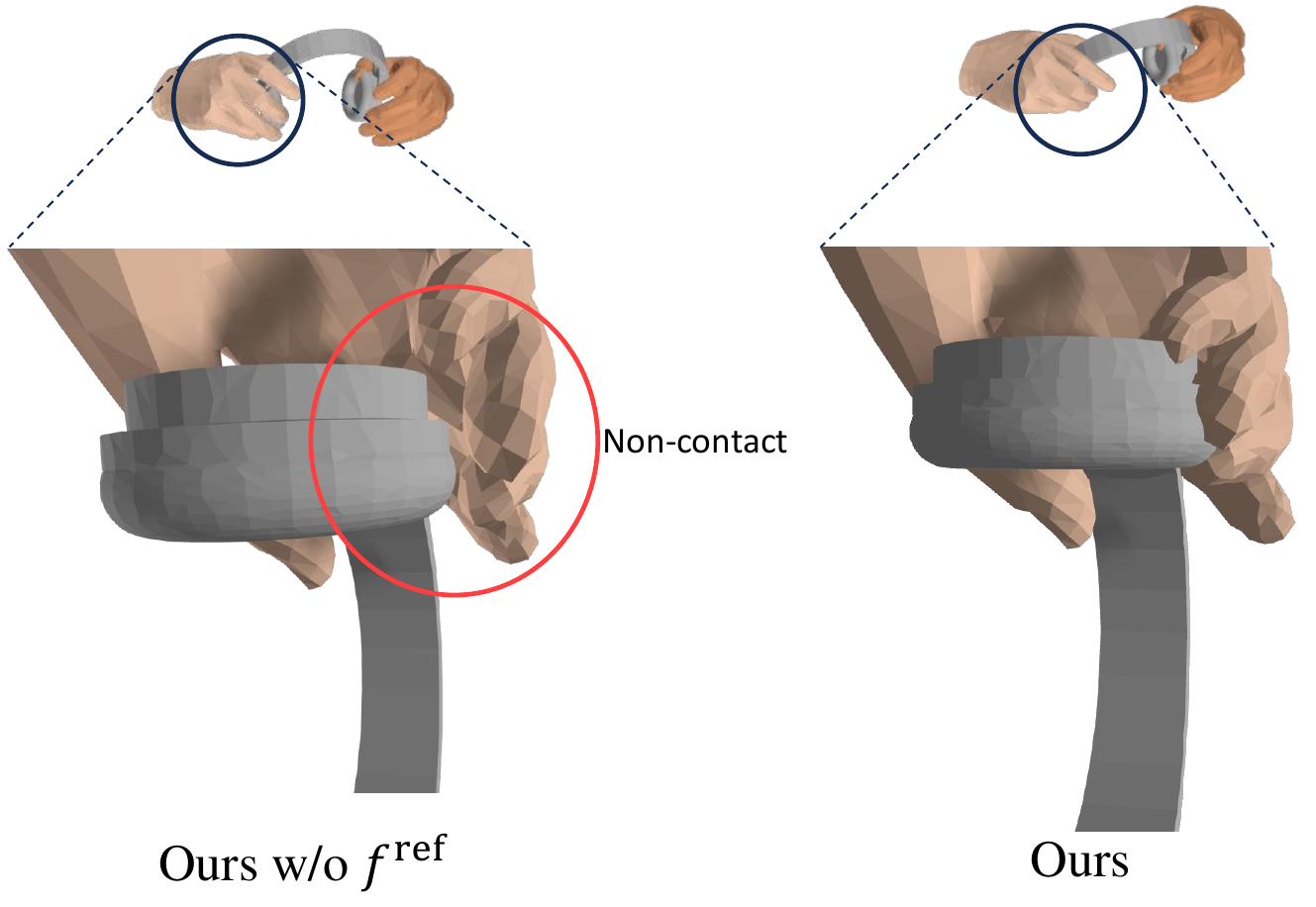}
\caption{Our hand refinement network refines the contacts.}
\label{fig:ablation refiner}
\end{figure}

\begin{figure}[t!]
\centering
\includegraphics[width=0.46\textwidth]{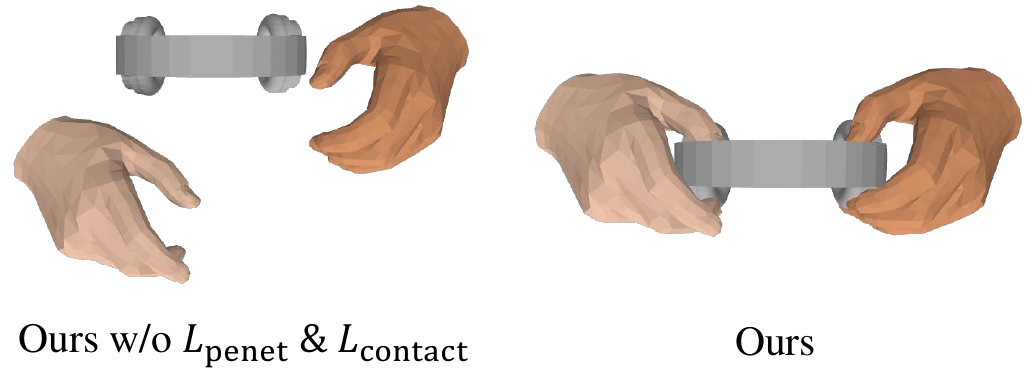}
\caption{Without the penetration loss and contact loss, the generated motions result in the hands and object not interacting.}
\label{fig:ablation refiner loss}
\end{figure}

\begin{figure}[t!]
\centering
\includegraphics[width=0.46\textwidth]{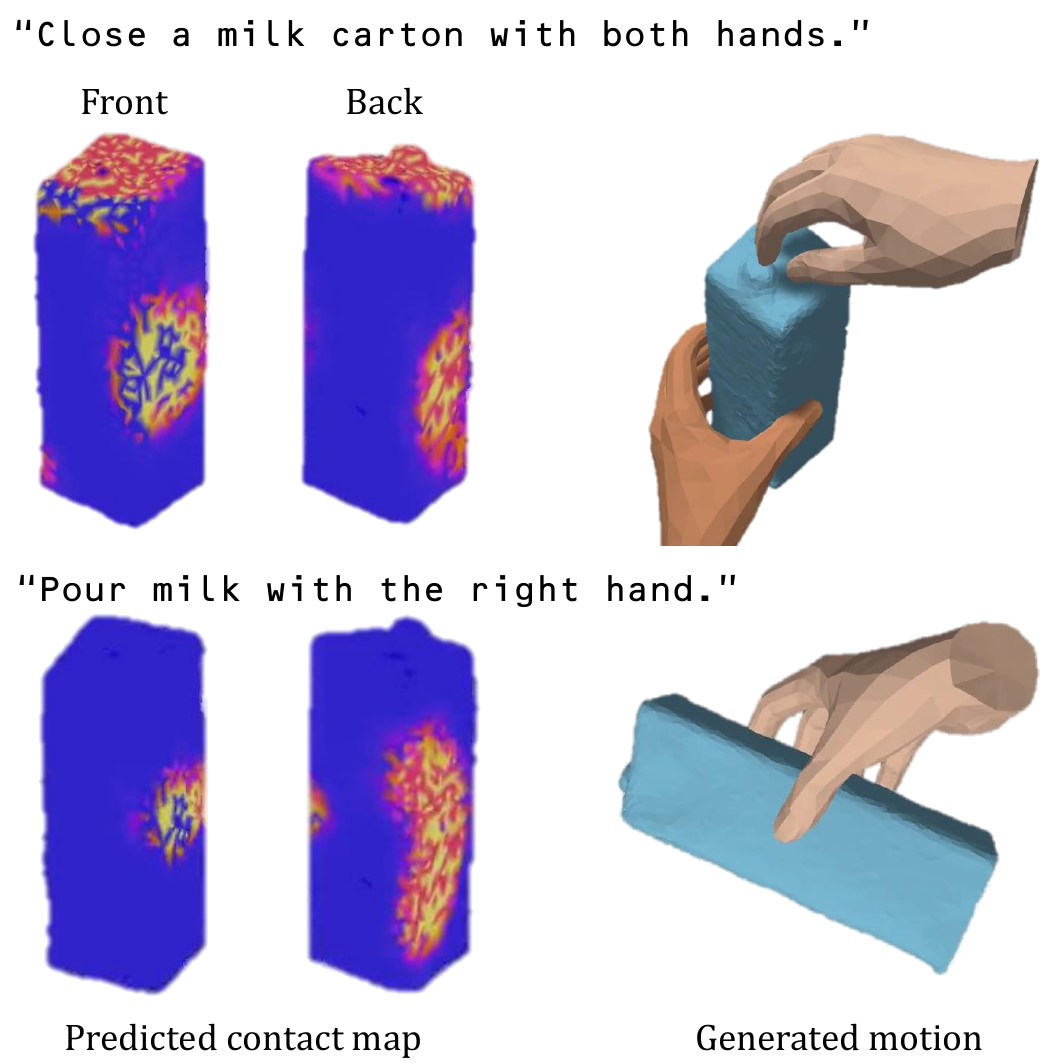}
\caption{We display the predicted contact map and the generated motion, focusing on their variations in response to different text prompts.}
\label{fig:contact map by text}
\end{figure}

\begin{figure}[t!]
\centering
\includegraphics[width=0.46\textwidth]{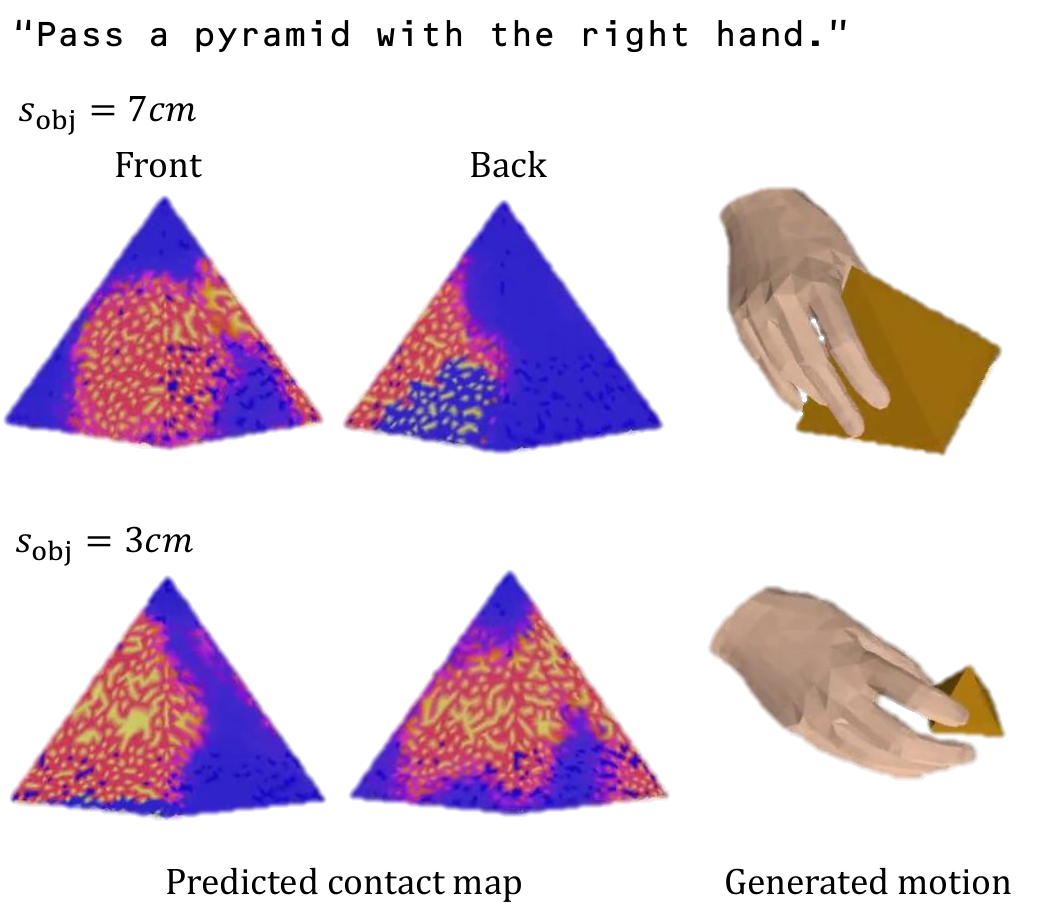}
\caption{We illustrate how the predicted contact map and generated motion vary across different object's scales.}
\label{fig:contact map by scale}
\end{figure}

\section{Qualitative results for ablation studies}
We present the qualitative results in Fig.~\ref{fig:ablation THOI}, to demonstrate the effectiveness of geometry losses (distance map loss $L_\text{dm}$ and relative orientation loss $L_\text{ro}$), conditions (contact map $\hat{\mbm}_\text{contact}$ and object's scale $s_\text{obj}$), and our proposed positional encodings (frame-wise and agent-wise positional encodings). In addition, we present the qualitative results related to hand refinement in Figs.~\ref{fig:ablation refiner} and \ref{fig:ablation refiner loss}, to show the effectiveness of refiner $f^\text{ref}$, and additional losses (penetration loss $L_\text{penet}$ and contact loss $L_\text{contact}$), respectively. In Figs.~\ref{fig:ablation THOI}, \ref{fig:ablation refiner}, and \ref{fig:ablation refiner loss}, the text prompt employed is \textit{``Use headphones with both hands."}.

We demonstrate the variant predicted contact maps and generated motions, as shown in Figs.~\ref{fig:contact map by text} and \ref{fig:contact map by scale}. The predicted contact maps accurately reflect the text prompts. The motions vary depending on the predicted contact maps and the prompts. In addition, the contact maps are predicted differently for different object's scales, influencing the number of fingers involved and the manner of grasping depending on the object's scale.

\section{Refiner design}
We compare our hand refinement network with the diffusion-based hand refinement network. It is counterintuitive to apply the diffusion-based approach for refining already generated motions by $f^\text{THOI}$. The diffusion model necessitates adding noise to the already generated motions through a diffusion forward process, and then denoising them via a backward process. Given this inefficiency, we propose a refiner that does not rely on the diffusion-based method. Additionally, in terms of performance, our refiner $f^\text{ref}$ demonstrates superior results as shown in Tab.~\ref{tab:hand refiner design}.

\begin{table}[t!]
    \centering
    \caption{Comparative physical realism scores for the different refiner designs.}
    \begin{tabular}{lc}
    \hline
    Method & Physical realism \\
    \hline
    Diffusion-based hand refiner     & 0.1682 $\pm$ 0.0006 \\
    Ours     & \textbf{0.8839 $\pm$ 0.0005} \\
    \hline
    \end{tabular}
    \label{tab:hand refiner design}
\end{table}

\section{Effect of refiner}
\begin{figure}[t!]
\centering
\includegraphics[width=0.46\textwidth]{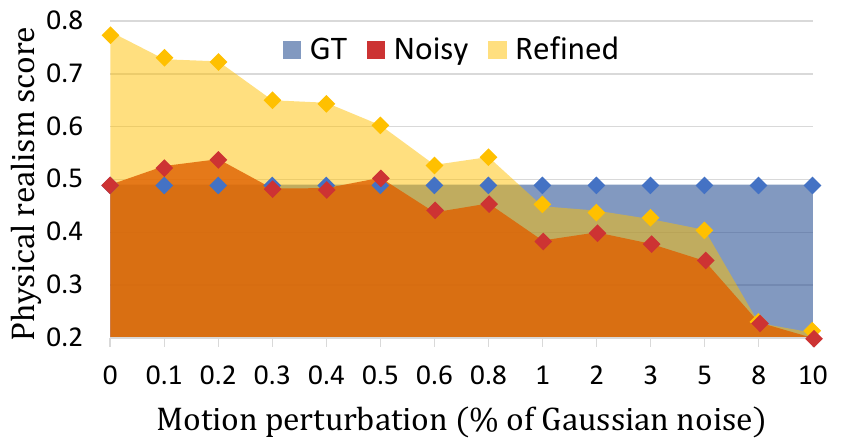}
\caption{The physical realism score with ground-truth (GT), noisy, and refined motion parameters. Synthetically created noisy motion parameters are produced by incorporating the function of Gaussian noise, which distorts the ground-truth motion parameters.}
\label{fig:refiner graph}
\end{figure}

We demonstrate the effectiveness of our hand refinement network $f^\text{ref}$ as shown in Fig.~\ref{fig:refiner graph}. The physical realism score is highly increased by our hand refinement network even it outperforms that of ground-truth motion. 

\section{Inference speed}
We measure the time required to generate 150 frames of hand-object interaction using an RTX 4090, with the detailed results presented in Tab.~\ref{tab:inference time}. Notably, our method demonstrates faster performance compared to the diffusion-based method MDM~\cite{tevet2023human}.

\begin{table}[t!]
    \centering
    \caption{Inference speed.}
    \begin{tabular}{clc}
    \hline
    \multicolumn{2}{l}{Method} & Time \\
    \hline
    MDM~\cite{tevet2023human} & & 28.5s \\
    IMoS~\cite{ghosh2023imos} & & 101s \\
    \hline
    \multirow{3}{*}{Ours} & Contact map $f^\text{contact}$ & 0.011s \\
     & Text2HOI $f^\text{THOI}$ & 4.9s \\
     & Refinement $f^\text{ref}$ & 0.013s \\
     \hline
    \end{tabular}
    \label{tab:inference time}
\end{table}

\section{Notations}
We summarize notations used in main paper and supplementary material in Tab.~\ref{tab:notations}.

\begin{table}[t!]
    \centering
    \caption{Notations.}
    \begin{tabular}{l|l}
    \hline
    Symbol & Definition \\
    \hline
    $\mbT$ & Text prompt \\
    $f^\text{CLIP}(\mbT)$ & Text features \\
    $\mbM_\text{obj}$ & Canonical object mesh \\
    $\mbH^*$ & Hand type (left, right, both) \\
    $\mbx$     &  Motion \\
    $\mbx^l$     & Motion at $l$-th frame  \\
    $\mbx_{t}$     & Noised motion at $t$-th diffusion time-step\\
    $\mbx_{0}$     & Clean motion \\
    $\mbx_{\text{lhand}}$     & Left hand motion \\
    $\mbx_{\text{rhand}}$     & Right hand motion \\
    $\mbx_{\text{hand}}$     & Hand motion \\
    $\mbx_{\text{obj}}$     & Object motion \\
    $T$     & The number of diffusion steps \\
    $L$     & Motion length \\
    $L_\text{max}$     & Maximum motion length (=150) \\
    $s_\text{obj}$ & Object's scale \\
    $\mbP$     & Object point cloud \\
    $\mbP_\text{def}$     & Deformed object point cloud \\
    $\mbF_\text{obj}$ & (Global) object features \\
    $\mbV$     & Hand vertices \\
    $V$     & The number of hand vertices (=778) \\
    $\mbJ$     & Hand joints \\
    $J$     & The number of hand joints (=21) \\
    $\mbm_\text{contact}$     & Contact map \\
    $\mbX$     & Embedded value (motion, condition) \\
    $\hat{\cdot}$ & Estimated value (output) \\
    $\Tilde{\cdot}$ & Refined value (output) \\
    $f$ & Network \\
    \hline
    \end{tabular}
    \label{tab:notations}
\end{table}

\begin{figure*}[t!]
\centering
\includegraphics[width=0.99\textwidth]{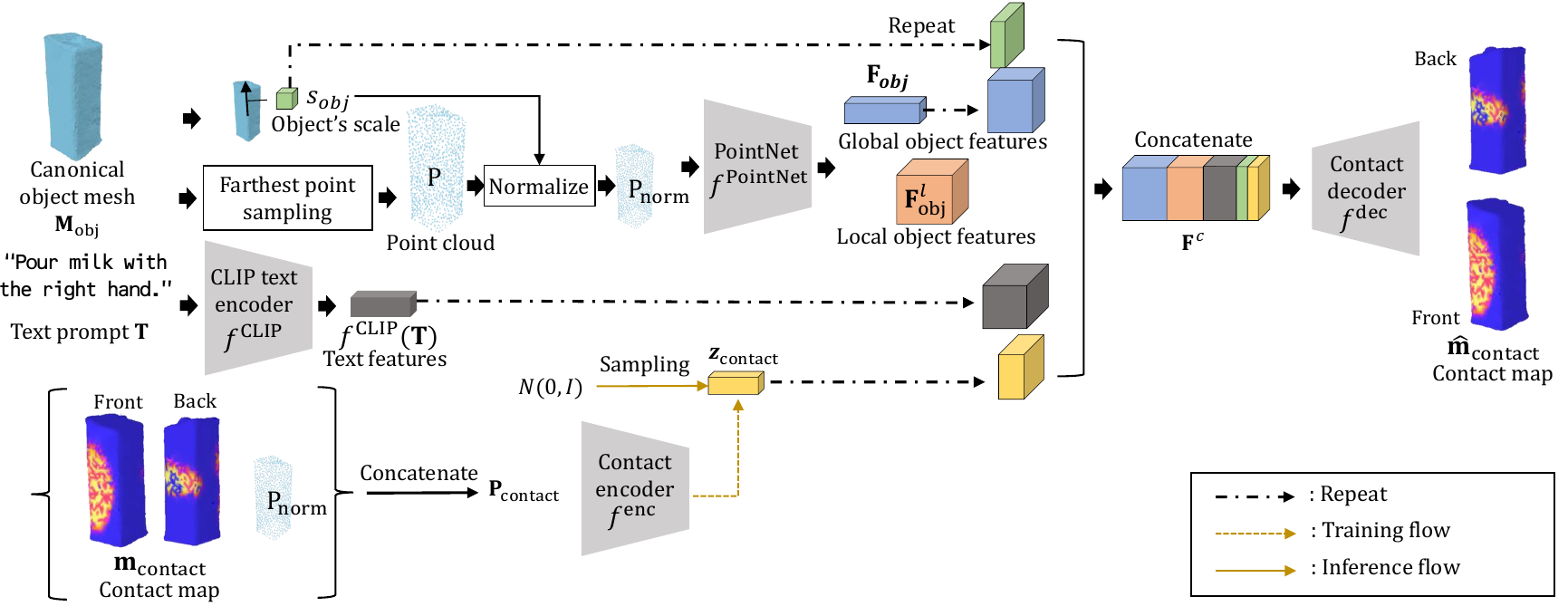}
\caption{Predicting the contact map consists of 4 steps. 1) From a canonical object mesh $\mbM_\text{obj}$, we compute the object's scale $s_\text{obj}$. Then, we sample the point cloud $\mbP$ using the farthest point sampling algorithm, and normalize $\mbP$ to $\mbP_\text{norm}$ by dividing it with $s_\text{obj}$. The PointNet $f^\text{PointNet}$ receives  $\mbP_\text{norm}$ as input, to extract the global object features $\mbF_\text{obj}$ and local object features $\mbF^l_\text{obj}$. 2) From a text prompt $\mbT$, CLIP text encoder $f^\text{CLIP}$ extracts the text features $f^\text{CLIP}(\mbT)$. 3) At training time, we extract the vector $\mbz_\text{contact}$ using the contact encoder $f^\text{enc}$ from $\mbP_\text{contact}$ that is concatenated with the ground-truth contact map $\mbm_\text{contact}$ and the normalized point cloud $\mbP_\text{norm}$. At inference time, we sample $\mbz_\text{contact}$ from Gaussian distribution. 4) We concatenate the diverse features $\mbF_\text{obj}$, $\mbF^l_\text{obj}$, $f^\text{CLIP}(\mbT)$, $s_\text{obj}$, and $\mbz_\text{contact}$, to produce $\mbF^c$. Finally, the contact decoder $f^\text{dec}$ predicts the contact map $\hat{\mbm}_\text{contact}$ from $\mbF^c$.}
\label{fig:contact map module details}
\end{figure*}

\section{Network}
Several networks are involved in our framework: contact prediction network $f^\text{contact}$, text-to-3D hand-object interaction generator $f^\text{THOI}$, and hand refinement network $f^\text{ref}$.

\noindent \textbf{Contact prediction network.} We predict the contact map $\mbm_\text{contact}$ from a text prompt $\mbT$ and a canonical object mesh $\mbM_\text{obj}$. The contact map prediction procedure involves four steps: 1) extracting object features from the object; 2) extracting text features from the text prompt; 3) sampling a Gaussian random noise vector; and 4) predicting the contact map, as illustrated in Fig.~\ref{fig:contact map module details}.

We first compute an object's scale $s_\text{obj} \in \mathbb{R}^1$ that represents the maximum distance from the center of object mesh $\mbM_\text{obj}$ to its vertices. Then, we sample $N$-point cloud $\mbP \in \mathbb{R}^{N \times 3}$ from the vertices of $\mbM_\text{obj}$ ($N=1,024$) using the farthest point sampling (FPS) algorithm~\cite{qi2017pointnet}. Subsequently, $\mbP$ is normalized to $\mbP_\text{norm}$ by dividing it with $s_\text{obj}$. We utilize PointNet $f^\text{PointNet}$~\cite{qi2017pointnet} to extract local object features $\mbF^l_\text{obj} \in \mathbb{R}^{N \times 64}$ and global object features $\mbF_\text{obj} \in \mathbb{R}^{1,024}$ from the normalized point cloud $\mbP_\text{norm}$. 

Second, we extract the text features $f^\text{CLIP}(\mbT) \in \mathbb{R}^{512}$ from the text prompt $\mbT$ using CLIP text encoder $f^\text{CLIP}$~\cite{radford2021learning}. 

Third, at training time, the vector $\mbz_\text{contact} \in \mathbb{R}^{64}$ is encoded from the concatenated input $\mbP_\text{contact} \in \mathbb{R}^{N \times 4}$ with $\mbP_\text{norm} \in \mathbb{R}^{N \times 3}$ and the ground-truth contact map $\mbm_\text{contact} \in \mathbb{R}^{N \times 1}$ using the contact encoder $f^\text{enc}$. At inference time, $\mbz_\text{contact}$ is sampled from Gaussian distribution. 

Fourth, we concatenate $\mbF_\text{obj}$, $\mbF^l_\text{obj}$, $s_\text{obj}$, $f^\text{CLIP}(\mbT)$, and $\mbz_\text{contact}$. Since they have varying feature dimensions, we duplicate these features $N$ times except $\mbF^l_\text{obj}$, to align the dimension shape: $\text{Repeat}(\mbF_\text{obj}): \mathbb{R}^{1,024} \rightarrow \mathbb{R}^{N \times 1,024}$, $\text{Repeat}(s_\text{obj}): \mathbb{R}^{1} \rightarrow \mathbb{R}^{N \times 1}$, $\text{Repeat}(f^\text{CLIP}(\mbT)): \mathbb{R}^{512} \rightarrow \mathbb{R}^{N \times 512}$, and $\text{Repeat}(\mbz_\text{contact}): \mathbb{R}^{64} \rightarrow \mathbb{R}^{N \times 64}$. Finally, the concatenated features $\mbF^{c} \in \mathbb{R}^{N \times 1,665}$ are fed to the contact decoder $f^\text{dec}$ to predict the contact map $\hat{\mbm}_\text{contact} \in \mathbb{R}^{N \times 1}$. The architectures of $f^\text{PointNet}$, $f^\text{enc}$, and $f^\text{dec}$ are detailed in Tabs.~\ref{tab:architecture contact map module pointnet}, \ref{tab:architecture contact map module encoder}, and \ref{tab:architecture contact map module decoder}, respectively.

\begin{table}[t]
    \centering
    \caption{Architecture of PointNet $f^\text{PointNet}$. $N$ represents the number of points of point cloud. $k$ denotes the kernel size. BN denotes a batch normalization. $I$ represents an identity matrix. $x$ denotes the output of the previous layer.}
    \resizebox{0.49\textwidth}{!}{
    \begin{tabular}{c|c|c}
        \hline
        Layer & Operation & Output Features \\
        \hline
        input & $\mbP_\text{norm}$ & $N \times 3$ \\
        transpose & transpose & $3 \times N$ \\
        STN-conv1 ($k$=1) & Conv 1D + BN + ReLU & $64 \times N$ \\
        STN-conv2 ($k$=1) & Conv 1D + BN + ReLU & $128 \times N$ \\
        STN-conv3 ($k$=1) & Conv 1D + BN + ReLU & $1,024 \times N$ \\
        max (axis=1) & max & $1,024$ \\
        STN-fc1 & Linear + BN + ReLU & $512$ \\
        STN-fc2 & Linear + BN + ReLU & $256$ \\
        STN-fc3 & Linear & $9$ \\
        reshape & reshape & $3 \times 3$ \\
        identity & $x = x + I$ & $3 \times 3$ \\
        multiplication & $\mbP_\text{norm} \times x$ & $N \times 3$ \\
        transpose & transpose & $3 \times N$ \\
        conv1 ($k$=1) & Conv 1D + BN + ReLU & $64 \times N$ \\
        transpose & transpose & $N \times 64$ \\
        assign & $\mbF^l_\text{obj} = x$ & $N \times 64$ \\
        transpose & transpose & $64 \times N$ \\
        conv2 ($k$=1) & Conv 1D + BN + ReLU & $128 \times N$ \\
        conv3 ($k$=1) & Conv 1D + BN & $1,024 \times N$ \\
        max (axis=1) & max & $1,024$ \\
        assign & $\mbF_\text{obj} = x$ & $1,024$ \\
        \hline
    \end{tabular}}
    \label{tab:architecture contact map module pointnet}
    \vspace{-5mm}
\end{table}

\begin{table}[t]
    \centering
    \caption{Architecture of contact encoder $f^\text{enc}$. $N$ represents the number of points of point cloud. $k$ denotes the kernel size. BN denotes the batch normalization. $x$ denotes the output of previous layer.}
    \resizebox{0.49\textwidth}{!}{
    \begin{tabular}{c|c|c}
        \hline
        Layer & Operation & Output Features \\
        \hline
        input & Concatenate($\mbP_\text{norm}, \mbm_\text{contact}$) & $N \times 4$ \\
        transpose & transpose & $4 \times N$ \\
        STN-conv1 ($k$=1) & Conv 1D + BN + ReLU & $64 \times N$ \\
        STN-conv2 ($k$=1) & Conv 1D + BN + ReLU & $128 \times N$ \\
        STN-conv3 ($k$=1) & Conv 1D + BN + ReLU & $1,024 \times N$ \\
        max (axis=1) & max & $1,024$ \\
        STN-fc1 & Linear + BN + ReLU & $512$ \\
        STN-fc2 & Linear + BN + ReLU & $256$ \\
        STN-fc3 & Linear & $16$ \\
        reshape & reshape & $4\times4$ \\
        identity & $x = x + I$ & $4 \times 4$ \\
        multiplication & $\mbP_\text{norm} \times x$  & $N \times 4$ \\
        transpose & transpose & $4 \times N$ \\
        conv1 ($k$=1) & Conv 1D + BN + ReLU & $64 \times N$ \\
        conv2 ($k$=1) & Conv 1D + BN + ReLU & $128 \times N$ \\
        conv3 ($k$=1) & Conv 1D + BN & $1,024 \times N$ \\
        max (axis=1) & max & $1,024$ \\
        fc1 & Linear + BN + ReLU & $512$ \\
        fc2 & Linear + BN + ReLU & $256$ \\
        fc3 & Linear & $128$ \\
        \hdashline
        \multirow{2}{*}{split} & mean $\mbz_{\mu}$ & $64$ \\
         & variance $\mbz_{\sigma^2}$ & $64$ \\
        \hdashline
        reparameterize & $\mbz_\text{contact}$ & $64$ \\
        \hline
    \end{tabular}}
    \label{tab:architecture contact map module encoder}
\end{table}

\begin{table}[t]
    \centering
    \caption{Architecture of contact decoder $f^\text{dec}$. `slope' denotes the negative slope of LeakyReLU.}
    \resizebox{0.49\textwidth}{!}{
    \begin{tabular}{c|c|c}
        \hline
        Layer & Operation & Output Features \\
        \hline
        input & $\mbF^{c}$ & $N \times 1,665$ \\
        fc1 & Linear + LeakyReLU(slope=0.2) & $N \times 512$ \\
        fc2 & Linear + LeakyReLU(slope=0.2) & $N \times 256$ \\
        fc3 & Linear + LeakyReLU(slope=0.2) & $N \times 128$ \\
        fc4 & Linear & $N \times 1$ \\
        \hline
    \end{tabular}}
    \label{tab:architecture contact map module decoder}
\end{table}

\noindent \textbf{Text-to-3D hand-object interaction generator.} Our generator $f^\text{THOI}$ receives several inputs: time-step $t$, text features $f^\text{CLIP}(\mbT)$, object features $\mbF_\text{obj}$, contact map $\hat{\mbm}_\text{contact}$, object's scale $s_\text{obj}$ as condition, and noised motion $\mbx_t$ as input. Then, it outputs the denoised motions $\hat{\mbx}_0$. The structure of $f^\text{THOI}$ includes various layers: input embedding layers ($f^\text{in, lhand}$, $f^\text{in, rhand}$, $f^\text{in, obj}$), condition embedding layers ($f^\text{ts}$, $f^\text{text}$, $f^\text{obj}$), a Transformer encoder, and output embedding layers ($f^\text{out, lhand}$, $f^\text{out, rhand}$, $f^\text{out, obj}$). The specific architectures of the input and condition embedding layers (see Tab.~\ref{tab:architecture THOI f in}), Transformer encoder (see Tab.~\ref{tab:architecture THOI encoder}), and output embedding layers (see Tab.~\ref{tab:architecture THOI f out}) are detailed in their respective tables.

\begin{table}[t]
    \centering
    \caption{Architectures of input and condition embedding layers in text-to-3D hand-object interaction generator $f^\text{THOI}$. $\{ \}$ denotes a concatenation.}
    \renewcommand{\arraystretch}{1.2}
    \resizebox{0.49\textwidth}{!}{
    \begin{tabular}{c|c|c}
        \hline
        Layer & Operation & Output Features \\
        \hline
        input & $\mbx^l_{t, \text{lhand}}$ & $99$ \\
        $f^\text{in, lhand}$ & Linear & $512$ \\
        \hline
        input & $\mbx^l_{t, \text{rhand}}$ & $99$ \\
        $f^\text{in, rhand}$ & Linear & $512$ \\
        \hline
        input & $\mbx^l_{t, \text{obj}}$ & $10$ \\
        $f^\text{in, obj}$ & Linear & $512$ \\
        \hline
        \hline
        input & $t$ & scalar \\
        $f^\text{ts}$-pe & Positional encoding & $512$ \\
        $f^\text{ts}$-fc1 & Linear + SiLU & $512$ \\
        $f^\text{ts}$-fc2 & Linear& $512$ \\
        \hline
        input & $f^\text{CLIP}(\mbT)$ & $512$ \\
        $f^\text{text}$ & Linear & $512$ \\
        \hline
        input & $\{ \mbF_\text{obj}, \hat{\mbm}_\text{contact}, s_\text{obj} \}$ & $2049$ \\
        $f^\text{obj}$ & Linear & $512$ \\
        \hline
    \end{tabular}}
    \label{tab:architecture THOI f in}
\end{table}

\begin{table}[t]
    \centering
    \caption{Architecture of Transformer encoder in text-to-3D hand-object interaction generator $f^\text{THOI}$. LN denotes a layer normalization. n=8 denotes that the layer repeat 8 times. $x$ denotes the output of previous layer. $\hat{L}$ represenst the estimated motion length.}
    \renewcommand{\arraystretch}{1.5}
    \resizebox{0.49\textwidth}{!}{
    \begin{tabular}{c|c|c|c}
        \hline
        \multicolumn{2}{c|}{Layer} & Operation & Output Features \\
        \hline
        \multicolumn{2}{c|}{input} & $\mbX_t$ & $(1+3\hat{L}) \times 512$ \\
        \hline
        \multirow{8}{*}{n=8} & Multi-Head Attention & Self attention & \multirow{2}{*}{$(1+3\hat{L}) \times 512$} \\
        & (h=4) & (SA) & \\
        \cdashline{2-4}
        & Residual & $\mbX_t$ = $\mbX_t$+SA & $(1+3\hat{L}) \times 512$ \\
        & Normalize1 & LN & $(1+3\hat{L}) \times 512$ \\
        & fc1 & Linear + GeLU & $(1+3\hat{L}) \times 1024$ \\
        & fc2 & Linear & $(1+3\hat{L}) \times 512$ \\
        & Residual & $\mbX_t$ = $\mbX_t+x$ & $(1+3\hat{L}) \times 512$ \\
        & Normalize2 & LN & $(1+3\hat{L}) \times 512$ \\
        \hline
    \end{tabular}}
    \label{tab:architecture THOI encoder}
\end{table}

\begin{table}[t]
    \centering
    \caption{Architectures of output embedding layers in text-to-3D hand-object interaction generator $f^\text{THOI}$.}
    \resizebox{0.4\textwidth}{!}{
    \begin{tabular}{c|c|c}
        \hline
        Layer & Operation & Output Features \\
        \hline
        input & $\mbX^l_{0, \text{lhand}}$ & $512$ \\
        $f^\text{out, lhand}$ & Linear & $99$ \\
        \hline
        input & $\mbX^l_{0, \text{rhand}}$ & $512$ \\
        $f^\text{out, rhand}$ & Linear & $99$ \\
        \hline
        input & $\mbX^l_{0, \text{obj}}$ & $512$ \\
        $f^\text{out, obj}$ & Linear & $10$ \\
        \hline
    \end{tabular}}
    \label{tab:architecture THOI f out}
\end{table}

\begin{table}[t]
    \centering
    \caption{Architecture of hand refinement network $f^\text{ref}$. $\mbX_\text{ref}$ denotes $\{ \mbX^l_\text{ref, lhand}, ~\mbX^l_\text{ref, rhand} \}^{\hat{L}}_{l=1}$.}
    \renewcommand{\arraystretch}{1.5}
    \resizebox{0.49\textwidth}{!}{
    \begin{tabular}{c|c|c|c}
        \hline
        \multicolumn{2}{c|}{Layer} & Operation & Output Features \\
        \hline
        \multicolumn{2}{c|}{Input} & $\mbx_\text{ref}$ & $2\hat{L} \times 2,273$ \\
        \multicolumn{2}{c|}{$f^\text{in, lhand}_\text{ref}$} & Linear & $\hat{L} \times 512$ \\
        \multicolumn{2}{c|}{$f^\text{in, rhand}_\text{ref}$} & Linear & $\hat{L} \times 512$ \\
        \hline
        \multicolumn{4}{c}{Transformer encoder}\\
        \hline
        \multirow{8}{*}{n=8} & Multi-Head Attention & Self attention & \multirow{2}{*}{$2\hat{L} \times 512$} \\
        & (h=4) & (SA) & \\
        \cdashline{2-4}
        & Residual & $\mbX_\text{ref}$ = $\mbX_\text{ref}$+SA & $2\hat{L} \times 512$ \\
        & Normalize1 & LN & $2\hat{L} \times 512$ \\
        & fc1 & Linear + GeLU & $2\hat{L} \times 1024$ \\
        & fc2 & Linear & $2\hat{L} \times 512$ \\
        & Residual & $\mbX_\text{ref}$ = $\mbX_\text{ref}+x$ & $2\hat{L} \times 512$ \\
        & Normalize2 & LN & $2\hat{L} \times 512$ \\
        \hline
        \multicolumn{2}{c|}{$f^\text{out, lhand}_\text{ref}$} & Linear & $\hat{L} \times 99$ \\
        \multicolumn{2}{c|}{$f^\text{out, rhand}_\text{ref}$} & Linear & $\hat{L} \times 99$ \\
        \hline
    \end{tabular}}
    \label{tab:architecture refiner}
\end{table}

\noindent \textbf{Hand refinement network.} Our hand refinement network $f^\text{ref}$ receives the generated hand motion $\hat{\mbx}_{0, \text{hand}} \in \mathbb{R}^{2\hat{L} \times 99}$ ($\hat{\mbx}_{0, \text{hand}}=\{\hat{\mbx}^l_{0, \text{lhand}} \in \mathbb{R}^{99}, \hat{\mbx}^l_{0, \text{rhand}} \in \mathbb{R}^{99}\}^{\hat{L}}_{l=1}$), hand joints $\hat{\mbJ}_\text{hand} \in \mathbb{R}^{2\hat{L} \times J \times 3}$ ($\hat{\mbJ}_\text{hand} = \{ \hat{\mbJ}^l_\text{lhand} \in \mathbb{R}^{J \times 3}, \hat{\mbJ}^l_\text{rhand} \in \mathbb{R}^{J \times 3} \}^{\hat{L}}_{l=1}$), contact map $\hat{\mbm}_\text{contact} \in \mathbb{R}^{N \times 1}$, deformed object's point cloud $\hat{\mbP}_\text{def} \in \mathbb{R}^{\hat{L} \times N \times 3}$, and distance-based attention map $\mbm_\text{att} \in \mathbb{R}^{2\hat{L} \times J \times 3}$ ($\mbm_\text{att}=\{\mbm^l_\text{att, left} \in \mathbb{R}^{J \times 3},\mbm^l_\text{att, right} \in \mathbb{R}^{J \times 3}\}^{\hat{L}}_{l=1}$) as input, where $\hat{L}$ is an estimated motion length. We concatenate them to use them as input. First, the hand joints are reshaped: $\hat{\mbJ}_\text{hand} \in \mathbb{R}^{2\hat{L} \times 3J}$. Second, the contact map is duplicated  $2\hat{L}$ times and reshaped: $\hat{\mbm}_\text{contact} \in \mathbb{R}^{2\hat{L} \times N}$. Third, the deformed object's point cloud is duplicated 2 times, and the computation of the norm is applied across the last dimension: $\hat{\mbP}_\text{def} \in \mathbb{R}^{2\hat{L} \times N}$. Fourth, the distance-based attention map is reshape: $\mbm_\text{att} \in \mathbb{R}^{2\hat{L}\times 3J}$. We concatenate $\hat{\mbx}_{0,\text{hand}}$, $\hat{\mbJ}_\text{hand}$, $\hat{\mbm}_\text{contact}$, $\hat{\mbP}_\text{def}$, and $\mbm_\text{att}$ to create the input $\mbx_\text{ref} = \{ \mbx^l_\text{ref, lhand}, \mbx^l_\text{ref, rhand} \}^{\hat{L}}_{l=1} \in \mathbb{R}^{2\hat{L} \times (99+3J+N+N+3J)}$. $(99+3J+N+N+3J)$ is 2,273, where $J=21$ and $N=1,024$. The hand inputs $\mbx^l_\text{ref, lhand}$, and $\mbx^l_\text{ref, rhand}$ are fed to hand input embedding layers $f^\text{in, lhand}_\text{ref}$ and $f^\text{in, rhand}_\text{ref}$, respectively, to obtain the embeddings $\mbX^l_\text{ref, lhand} \in \mathbb{R}^{512}$ and $\mbX^l_\text{ref, rhand} \in \mathbb{R}^{512}$. Then, they are applied the frame-wise and agent-wise positional encodings and masked using $\mbH^*$, and fed to Transformer encoder. Then, Transformer encoder outputs the refined embeddings $\Tilde{\mbX}^l_\text{lhand} \in \mathbb{R}^{99}$ and $\Tilde{\mbX}^l_\text{rhand} \in \mathbb{R}^{99}$. These embeddings are passed through $f^\text{out, lhand}_\text{ref}$ and $f^\text{out, rhand}_\text{ref}$ and converted to refined hand motions $\Tilde{\mbx}^l_\text{lhand}$ and $\Tilde{\mbx}^l_\text{rhand}$. The final refined hand motions is expressed as follows: $\Tilde{\mbx}_\text{hand}=\{ \Tilde{\mbx}^l_\text{lhand}, \Tilde{\mbx}^l_\text{rhand}\}^{\hat{L}}_{l=1}$. The architecture of $f^\text{ref}$ is detailed in Tab.~\ref{tab:architecture refiner}.

\section{Masking inputs, outputs, and losses}
Using the hand-type variable $\mbH^*$, we implement masking in three areas: 1) the inputs of the Transformer encoder, 2) the outputs of the Transformer decoder, and 3) the losses ($L_\text{dm}$, $L_\text{ro}$, $L_\text{penet}$, $L_\text{contact}$). The masking process depends on the representation of $\mbH^*$ as follows:
\begin{itemize}
    \item If $\mbH^*$ represents the `left hand', then the inputs, outputs, and losses pertaining to the `right hand' are masked.
    \item Conversely, if $\mbH^*$ represents the `right hand', the corresponding components for the `left hand' are masked.
    \item If $\mbH^*$ indicates `both hands', no masking is applied to the inputs, outputs, or losses.
\end{itemize}

The indicator functions $\mathds{1}_\text{left}$ and $\mathds{1}_\text{right}$ are defined according to $\mbH^*$ as follows:
\begin{equation}
\{ \mathds{1}_\text{left}, \mathds{1}_\text{right} \} = \begin{cases}
\{1, 0\}, & \text{if $\mbH^*$ is `left hand'} \\ 
\{0, 1\}, & \text{if $\mbH^*$ is `right hand'} \\
\{1, 1\}, & \text{if $\mbH^*$ is `both hands'}
\end{cases}
\end{equation}

Masking the inputs means that the attention mechanism is inhibited for those inputs. Masking the outputs results in the visualization of those outputs being blocked. Masking the losses implies that backpropagation for those losses is restricted.

\section{Implementation details for baselines}
We maintained the baselines' model architecture, training scheme, and inference process, and just adjusted the model's output dimension to obtain parameters for two hands and for object~\cite{guo2022generating,tevet2023human,ghosh2023imos}. We used their pre-estimated length if they had a length estimator; otherwise, we used their predefined fixed length. We employed hand-type selection for masking the hand input and hand output, following the approach of our method.

\section{How to use the articulation parameter}
The articulation parameter is generated for all objects, regardless of their type. However, its actual application depends on datasets using articulation indicator. For articulated objects in ARCTIC dataset, the indicator is set as true and the articulation is reflected. For rigid objects in H2O and GRAB datasets, the indicator is set as false and the articulation parameter essentially acts as a placeholder and is not applied.

\section{More qualitative results}
\begin{figure*}[ht]
    \centering
    \includegraphics[width=0.95\textwidth]{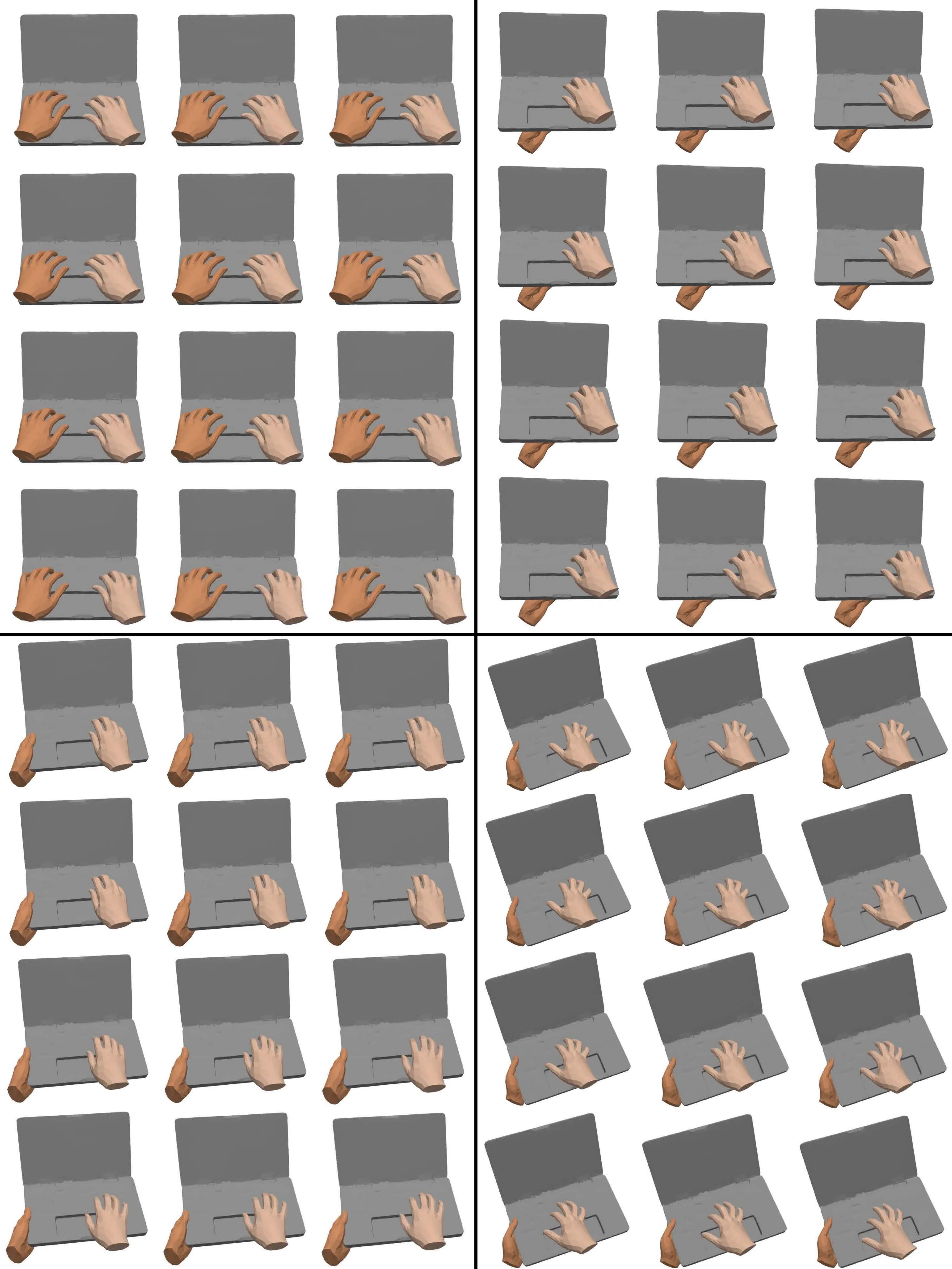}
    \caption{The diverse hand-object interactions from the same prompt \textit{`Type a laptop with both hands.'}. The sequence is from left to right and top to bottom.}
    \label{fig:diverse}
\end{figure*}

\begin{figure*}[ht]
    \centering
    \includegraphics[width=0.98\textwidth]{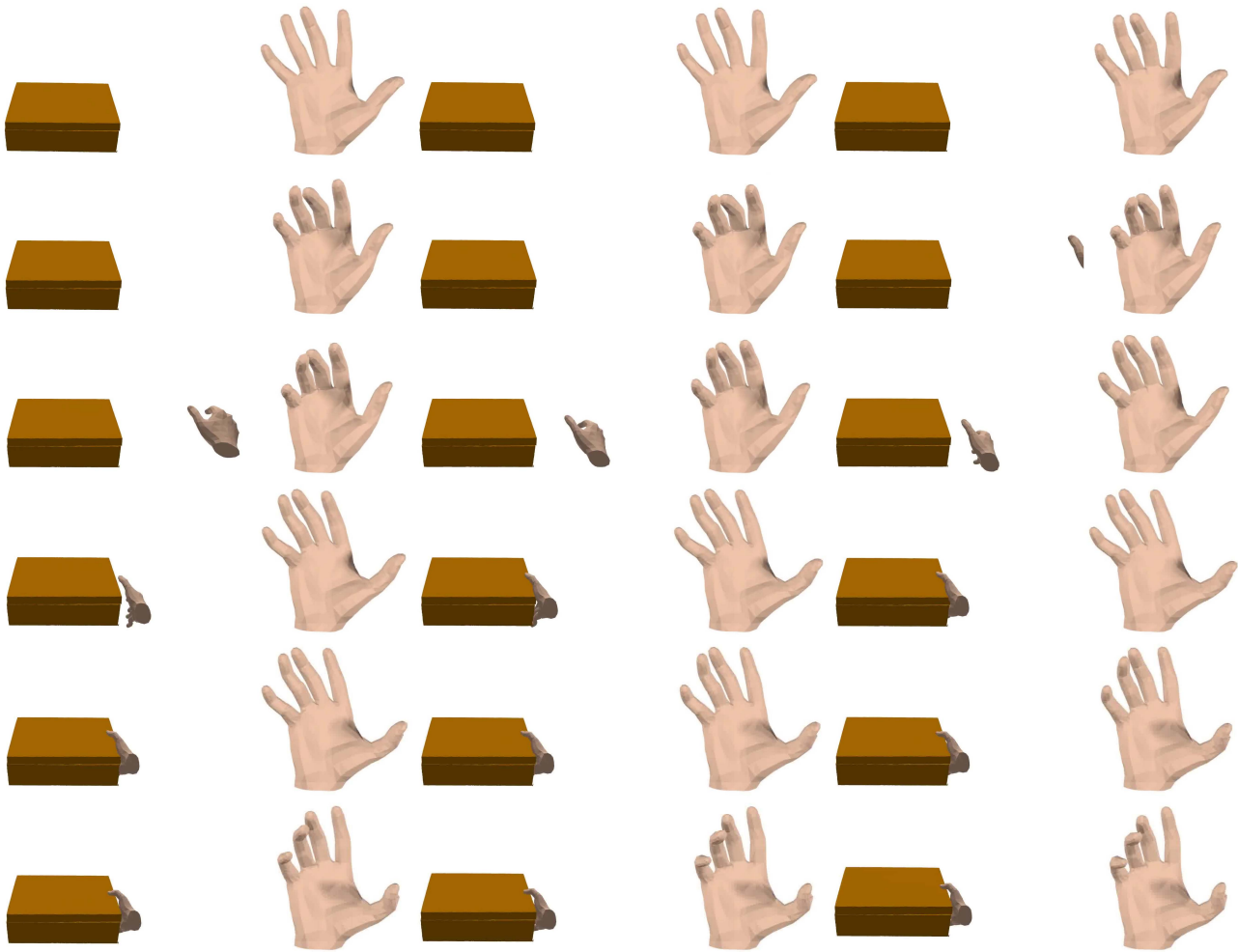}
    \caption{Generated motions (left) and hands in MANO canonical space (right) for the text prompt, \textit{`Grab a box with the right hand.'}. The sequence is from left to right and top to bottom.}
    \label{fig:canonical}
\end{figure*}

\begin{figure*}[ht]
    \centering
    \includegraphics[width=0.98\textwidth]{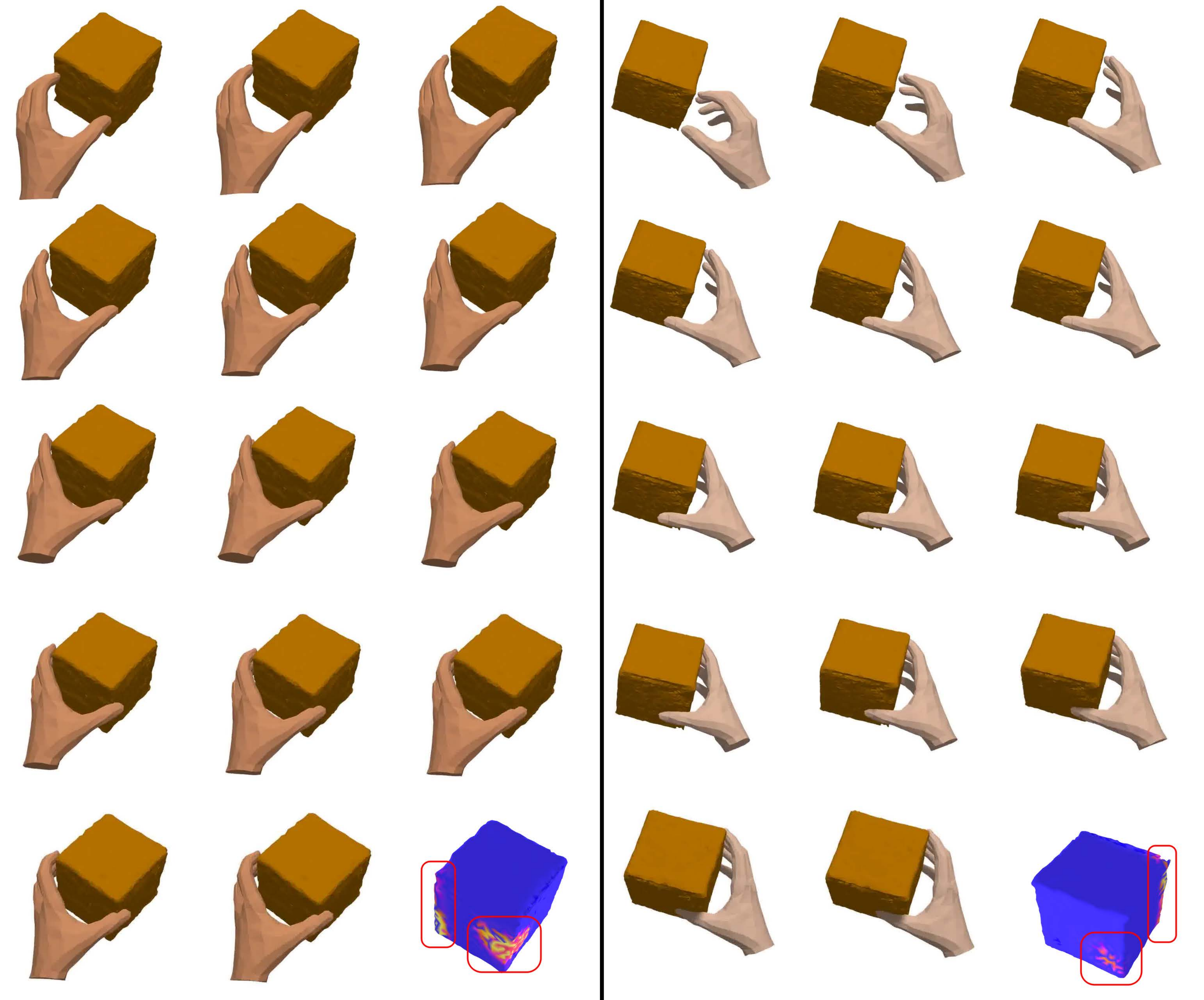}
    \caption{The different hand type results for the text prompts \textit{`Grasp a cappuccino with the left hand.'} and \textit{`Grasp a cappuccino with the right hand.'}. The sequence is from left to right and top to bottom.}
    \label{fig:hand_type}
\end{figure*}

Fig.~\ref{fig:diverse} shows the diverse hand-object interactions from the same prompt \textit{`Type a laptop with both hands.'}. Fig.~\ref{fig:canonical} demonstrates the hand pose change in canonical coordinate. Fig.~\ref{fig:hand_type} illustrates the varying results according to different hand types. Each figure illustrates a sequence of key frames extracted from a video, displayed in a grid format of M rows and N columns. The frames are organized to represent the temporal progression of the video from left to right and top to bottom, simulating the temporal order of the events depicted in the video. 

\section{Limitation.} 
Hand-object interacting motions are generated from the text prompt, considering the relative 3D location and contact between hands and an object; while we are missing forces between them, which may provide better physical understanding. Future works may need to consider such new aspects.

\end{document}